\documentclass[lettersize,journal]{IEEEtran}

\usepackage[pdftex]{graphicx}
\usepackage{amsmath} 
\usepackage{multirow}
\usepackage{booktabs}
\usepackage[ruled,vlined]{algorithm2e}
\usepackage{tikz}
\usepackage[caption=false,font=normalsize,labelfont=sf,textfont=sf]{subfig}
\usetikzlibrary{positioning, arrows.meta, shapes.geometric}
\usepackage[export]{adjustbox}
\usepackage[table]{xcolor} 
\usepackage{colortbl}
\usepackage{tabularx}
\usepackage{amssymb} 
\usepackage{pifont}   
\usepackage{stfloats}

\usepackage{amsfonts}
\usepackage{cite} 
\usepackage{hyperref}
\definecolor{lightblue}{RGB}{204, 231, 242} 
\begin{document}

\title{IPEC: Test-Time Incremental Prototype Enhancement Classifier for Few-Shot Learning}

\author{Wenwen Liao, Hang Ruan, Jianbo Yu, Xiaofeng Yang, Qingchao Jiang,~\IEEEmembership{Senior Member,~IEEE}, Xuefeng Yan,~\IEEEmembership{Member,~IEEE}
\thanks{This work is supported by the National Natural Science Foundation of China (Grant Number: 62203118) and the Fundamental Research Funds for the Central Universities (Grant Number: 2025SMECP012). 
 (\textit{Corresponding author: Jianbo Yu)}}
\thanks{Wenwen Liao and Hang Ruan are with the College of Intelligent Robotics and Advanced Manufacturing, Fudan University, Shanghai, 201203, China. Email: 24110860009@m.fudan.edu.cn, 23110860012@m.fudan.edu.cn.}
\thanks{Jianbo Yu and Xiaofeng Yang are with the School of Microelectronics, Fudan University, Shanghai, 201203, China. Email: jb\underline{\hspace{0.5em}}yu@fudan.edu.cn, xf\underline{\hspace{0.5em}}yang@fudan.edu.cn.}
\thanks{Qingchao Jiang and Xuefeng Yan are with the Key Laboratory of Advanced Control and Optimization for Chemical Processes of Ministry of Education, East China University of Science and Technology, Shanghai 200237, China. Email: qchjiang@ecust.edu.cn, xfyan@ecust.edu.cn.}}

\markboth{Journal of \LaTeX\ Class Files,~Vol.~14, No.~8, August~2015}%
{Shell \MakeLowercase{\textit{et al.}}: Bare Demo of IEEEtran.cls for IEEE Journals}

\maketitle

\begin{abstract}
Metric-based few-shot approaches have gained significant popularity due to their relatively straightforward implementation, high interpretability, and computational efficiency. However, stemming from the batch-independence assumption during testing, which prevents the model from leveraging valuable knowledge accumulated from previous batches. To address these challenges, we propose a novel test-time method called Incremental Prototype Enhancement Classifier (IPEC), a test-time method that optimizes prototype estimation by leveraging information from previous query samples. IPEC maintains a dynamic auxiliary set by selectively incorporating query samples that are classified with high confidence. To ensure sample quality, we design a robust dual-filtering mechanism that assesses each query sample based on both global prediction confidence and local discriminative ability. By aggregating this auxiliary set with the support set in subsequent tasks, IPEC builds progressively more stable and representative prototypes, effectively reducing its reliance on the initial support set. We ground this approach in a Bayesian interpretation, conceptualizing the support set as a prior and the auxiliary set as a data-driven posterior, which in turn motivates the design of a practical “warm-up and test" two-stage inference protocol. Extensive empirical results validate the superior performance of our proposed method across multiple few-shot classification tasks.
\end{abstract}

\begin{IEEEkeywords}
Few-Shot Learning, Test-Time Adaptation, Prototype Enhancement, Confidence-Based Selection.
\end{IEEEkeywords}

\IEEEpeerreviewmaketitle

\section{Introduction}

\begin{figure*}[ht]
  \centering
  \vspace{-3mm} 
  \includegraphics[width=1.0\linewidth]{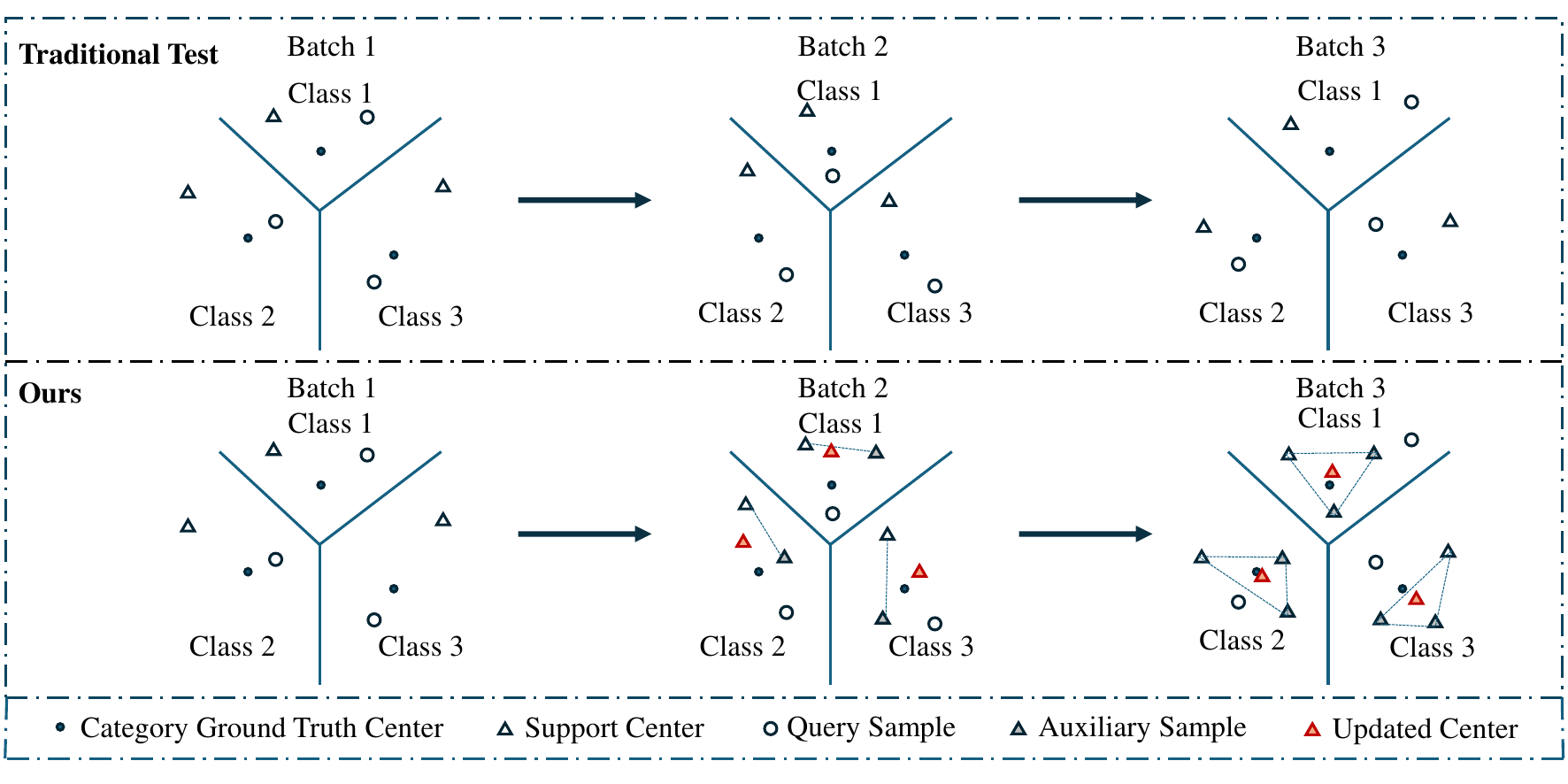} 
  \vspace{-5mm} 
  \caption{Comparison of Traditional Test and Ours test method.}
  \label{fig:tradition}
  \vspace{-5mm} 
\end{figure*}

\IEEEPARstart{I}{n} recent years, Few-Shot Learning (FSL) addresses the critical challenge of learning from scarce labeled data by enabling generalization to novel classes. Current research is dominated by following primary strategies: optimization-based methods that use meta-learning for rapid adaptation \cite{Finn2017, Ravi2017, Rusu2019, Baik2020}, and data augmentation methods that expand training sets with synthetic data \cite{Wang2018, Zhang2019, Wang2020}. Despite their utility, both approaches have significant limitations. Optimization-based methods often require extensive meta-training and generalize poorly to out-of-distribution tasks. Meanwhile, data augmentation's success is contingent on the quality of generated samples, which can be unreliable when data is extremely limited.

To address these issues, another effective approach is metric-based methods, which have attracted increasing attention due to their simplicity, interpretability, and efficiency. Metric-based methods aim to learn an embedding space where samples from the same category are close together, while samples from different categories are farther apart \cite{Fu2022}. Typically, this process involves projecting all input instances into vectors with a fixed dimensionality, with predictions for query samples made using methods such as nearest neighbor classifiers \cite{Vinyals2016}, robust category weights \cite{Tang2022}, parameterized metrics \cite{Sung2018}, or other task-specific distance metrics \cite{Liu2020}. Among them, Prototypical Networks (PN) \cite{Snell2017} are a prominent metric-based approach that computes a class prototype by averaging the embeddings of its support examples. Query samples are then classified based on their distance to these prototypes, and the method's simplicity and effectiveness have made it a foundational approach in the field \cite{Ji2020, Ding2020, Pahde2021, Gogoi2022}.

Although the aforementioned methods have demonstrated strong performance, they primarily rely on modifications made during the training phase \cite{yang2024adapting,zhao2024angular,yang2025hyperbolic,jing2025recursive}, leaving the potential for improvement during the model’s deployment or testing phase largely unexplored. A significant challenge in the test-time setting stems from the batch-independence assumption, which prevents the model from leveraging valuable information accumulated across batches during inference. To address this limitation, recent research in test-time adaptation (TTA) has focused on enhancing model generalization by adapting to distribution shifts or novel categories using unlabeled data available at test time. A variety of strategies have been proposed in the TTA literature. For instance, SHOT \cite{liang2020we} combined information maximization with pseudo-labeling to facilitate target domain adaptation. TENT \cite{wang2020tent} adapted models by updating the parameters of Batch Normalization layers via entropy minimization, while other approaches \cite{li2018adaptive, nado2020evaluating, schneider2020improving} dynamically estimated Batch Normalization statistics from incoming test data to better align with the target distribution. Although these methods have demonstrated impressive results across various domains, to the best of our knowledge, none of the existing TTA approaches have been adapted or applied to metric-based FSL.

This lack of exploration is particularly critical because metric-based FSL poses unique challenges at test time. As illustrated in the upper part of Fig. \ref{fig:tradition}, conventional testing procedures in metric-based FSL typically involve selecting a support set and a query set for each batch. The label of a query sample is then predicted based on its distance to class prototypes derived solely from the current support set. However, this approach treats each batch independently, meaning that the model’s decision-making process remains static after training. Such batch independence limits the system’s ability to leverage the valuable information embedded in abundant query samples encountered during inference, ultimately resulting in suboptimal performance—particularly in dynamic or evolving environments.

To overcome these challenges, and inspired by recent advances in test-time adaptation (TTA), a novel approach of test-time prototype adaptation is introduced into the FSL framework. This work presents an improved testing strategy, termed the Incremental Prototype Enhancement Classifier (IPEC), which progressively refines class representations during the test phase to ultimately enhance classification performance. In contrast to the conventional paradigm where support and query sets are treated independently, the proposed method (illustrated in the lower part of Fig. \ref{fig:tradition}) leverages confidently classified query samples from previous batches as auxiliary data to enhance prototype quality in subsequent ones. A dual-filtering mechanism is employed to select query samples based on both low entropy and a large local margin between the top-1 and top-2 predicted logits. As testing progresses, these selected samples accumulate in an auxiliary set, which is then combined with the current batch’s support set to compute updated and more reliable prototypes.

The principal contributions of this paper are as follows:
\begin{enumerate}
\item The proposal of IPEC, a novel few-shot testing method that incrementally enhances class prototypes using data from preceding batches. By establishing a dynamic, memory-augmented mechanism, IPEC breaks the static, batch-independent nature of conventional few-shot testing, enabling robust adaptation to the data stream encountered during inference.
\item The design of a dual-discrimination mechanism based on both global and local confidence for selecting high-quality samples. This is complemented by a straightforward filtering strategy to enhance the auxiliary set's quality and prevent error accumulation, ensuring the long-term integrity of the set by actively purging potentially mislabeled samples.
\item The modeling of the auxiliary set as a posterior sample set, which includes a theoretical proof of the asymptotic convergence of the prototype estimation. This analysis underpins a proposed two-stage inference protocol designed to further improve test-phase accuracy, motivating the practical and efficient “warm-up and test" design.
\item Comprehensive experimental validation on multiple datasets, demonstrating that IPEC significantly outperforms baseline methods and several state-of-the-art (SOTA) approaches. The results also show a progressive improvement in accuracy as test batches increase, highlighting the method's incremental learning capability.
\end{enumerate}

This paper is structured as follows. Section~\ref{sec:preliminaries} provides the necessary background. Section~\ref{sec:method} then details our proposed IPEC method. Comprehensive experimental results are reported in Section~\ref{sec:experiments}, and Section~-ref{sec:conclusion} concludes the paper with a discussion on future work.

\section{Preliminaries}
\label{sec:preliminaries}

\subsection{Problem Formulation}

The FSL task is formulated within the meta-learning paradigm. In this setting, the model is trained through a series of batches, each designed to mimic a few-shot classification scenario. Specifically, each batch consists of two subsets: a support set and a query set. The support set contains a small number of labeled examples from $N$ classes, which are used to construct a task-specific classifier. The query set, sampled from the same $N$ classes, is used to evaluate the classifier's performance. During training, the model parameters are updated using the loss computed on the query set, while during testing, the accuracy on the query set serves as the evaluation metric.

Let $\mathcal{D}_{\text{base}} = \{(\boldsymbol{x}, y) \mid y \in \mathcal{C}_{\text{base}} \}$ and $\mathcal{D}_{\text{novel}} = \{(\boldsymbol{x}, y) \mid y \in \mathcal{C}_{\text{novel}} \}$ denote the meta-training and meta-testing datasets, respectively. The sets of base classes $\mathcal{C}_{\text{base}}$ and novel classes $\mathcal{C}_{\text{novel}}$ are disjoint, i.e., $\mathcal{C}_{\text{base}} \cap \mathcal{C}_{\text{novel}} = \emptyset$. The objective of FSL is to train a model on $\mathcal{D}_{\text{base}}$ that can generalize effectively to $\mathcal{D}_{\text{novel}}$, where only a few labeled instances per class are available (e.g., 1-shot or 5-shot).

Following previous works~\cite{ Sung2018,Snell2017}, we adopt the $N$-way $K$-shot meta-learning framework. In each batch $\mathcal{T}$, we randomly sample $N$ classes from $\mathcal{C}_{\text{base}}$ (for training) or from $\mathcal{C}_{\text{novel}}$ (for testing), and select $K$ labeled examples per class to construct the support set $\mathcal{S} = \{(\boldsymbol{x}^{s}, y^{s})\}$. In addition, we sample $M$ query examples per class to form the query set $\mathcal{Q} = \{(\boldsymbol{x}^{q}, y^{q})\}$, ensuring that $\mathcal{S} \cap \mathcal{Q} = \emptyset$. Thus, each batch is defined as $\mathcal{T} = \{\mathcal{S}, \mathcal{Q}\}$, consisting of $N \times K$ support examples and $N \times M$ query examples. Unlike some methods (e.g.,~\cite{Snell2017}) that use different values of $K$ for training and testing, we maintain the same number of shots during both phases to ensure consistency and stable performance.

\subsection{Prototypical Networks}
\label{pn}

Our method is an improved variant based on PN \cite{Snell2017}, a representative metric-based FSL approach. PN learns an embedding space in which classification is performed by computing distances between embedded query samples and class prototypes—each prototype being the mean vector of embedded support examples from the corresponding class. The key idea is to represent each class by the mean of its embedded support examples (the prototype) and assign query samples to the class whose prototype is closest in the embedding space.

Let $f: \mathbb{R}^{n \times n} \rightarrow \mathbb{R}^d$ be the embedding function, which maps input features $\boldsymbol{x} \in \mathbb{R}^{n \times n}$ to an $d$-dimensional embedding space. Given an $N$-way $K$-shot support set $\mathcal{S} = \{(\boldsymbol{x}^{s}_i, y^{s}_i)\}_{i=1}^{N \times K}$, the prototype $\boldsymbol{p}_c$ for each class $c \in \{1, \dots, N\}$ is computed as the mean vector of the embedded support samples of that class:

\begin{equation}
    \boldsymbol{p}_c = \frac{1}{K} \sum_{(\boldsymbol{x}^{s}_i, y^{s}_i) \in \mathcal{S}_c} f(\boldsymbol{x}^{s}_i),
\label{p}
\end{equation}

\noindent where $\mathcal{S}_c$ denotes the subset of the support set belonging to class $c$.

For a query sample $\boldsymbol{x}^q$, the model computes the distance between its embedding and each class prototype. In the standard formulation, the Euclidean distance is commonly used:

\begin{equation}
    \text{Logit}_c^{\text{Euclid}}(\boldsymbol{x}^q, \boldsymbol{p}_c) = \left\|f(\boldsymbol{x}^q) - \boldsymbol{p}_c\right\|_2^2,
    \label{distance_euclid}
\end{equation}

Alternatively, cosine similarity is also employed in some PN variants or metric-based models:

\begin{equation}
    \text{Logit}_c^{\text{cos}}(\boldsymbol{x}^q, \boldsymbol{p}_c) =\cos\left(f(\boldsymbol{x}^q), \boldsymbol{p}_c\right)= \frac{f(\boldsymbol{x}^q) \cdot \boldsymbol{p}_c}{\|f(\boldsymbol{x}^q)\| \, \|\boldsymbol{p}_c\|},
    \label{distance_cosine}
\end{equation}

\noindent where \( \cos(\cdot, \cdot) \) denotes cosine similarity. 

The probability that the query belongs to class $c$ is obtained via a softmax over the negative distances:

\begin{equation}
    p(y = c \mid \boldsymbol{x}^q) = \frac{\exp(-d(\boldsymbol{x}^q, \boldsymbol{p}_c))}{\sum_{j=1}^{N} \exp(-d(\boldsymbol{x}^q, \boldsymbol{p}_j))}.
\end{equation}

The model is trained by minimizing the negative log-probability of the true class over all query samples in the query set $\mathcal{Q} = \{(\boldsymbol{x}^q_i, y^q_i)\}_{i=1}^{N \times M}$:

\begin{equation}
    \mathcal{L} = - \sum_{(\boldsymbol{x}^q, y^q) \in \mathcal{Q}} \log p(y = y^q \mid \boldsymbol{x}^q).
\end{equation}

During testing, the same procedure is applied to unseen classes from $\mathcal{C}_{\text{novel}}$, using their few labeled examples to compute class prototypes.

Compared to other meta-learning approaches, PN are simple yet effective, and they avoid the need for learning a meta-learner by relying solely on distance-based inference in an embedding space optimized for intra-class compactness and inter-class separability.

\begin{figure*}[ht]
  \centering
  \vspace{-3mm} 
  \includegraphics[width=1.0\linewidth]{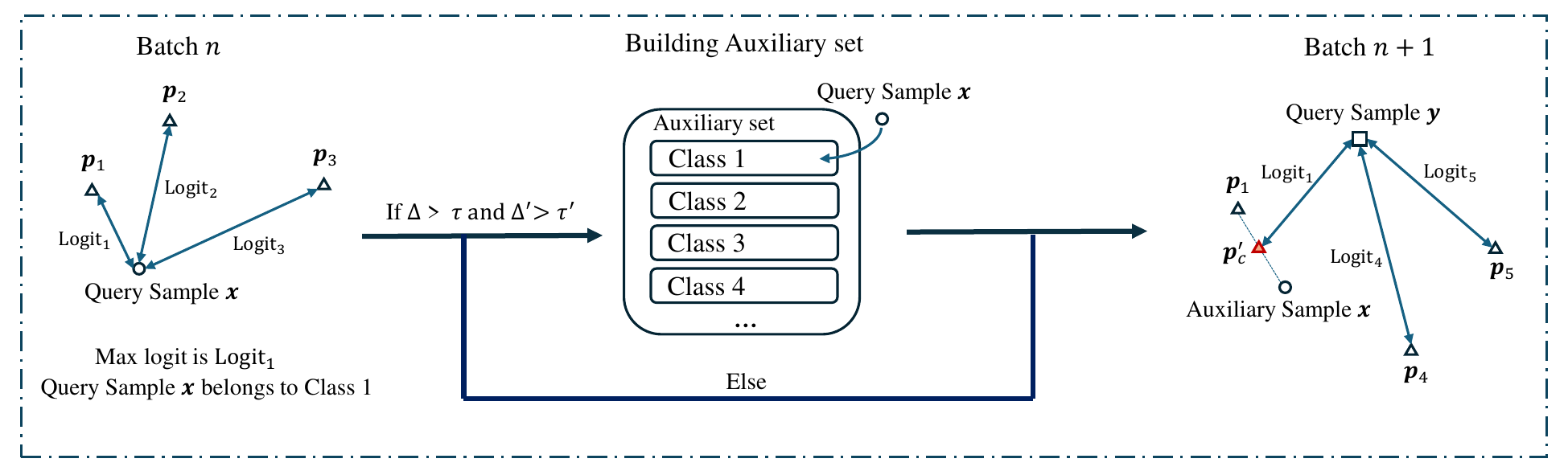} 
  \vspace{-5mm} 
  \caption{Pipeline of our proposed method.}
  \label{fig:pipeline}
  \vspace{-5mm} 
\end{figure*}

\section{Methodology}
\label{sec:method}

\subsection{Confidence-Based Auxiliary Set Construction}
\label{set construction}

The conventional approach computes class prototypes in feature space and applies a simple distance function (e.g., Euclidean distance or cosine similarity as in Eq. (\ref{distance_euclid}) and Eq. (\ref{distance_cosine})) between the query and each prototype. However, such methods neglect the underlying distributional structure of each class and may suffer from unreliable distance estimation, particularly in few-shot scenarios where only a handful of support samples are available.

In this work, a novel testing mechanism is proposed, built upon a probabilistic extension of PN. The conventional assumption of independence between test batches is broken by leveraging high-quality samples from previous batches to enhance prototype estimation. The key idea of IPEC lies in constructing an auxiliary set \( \mathcal{A} = \bigcup_c \mathcal{A}_c \), which selectively incorporates confidently predicted query samples. Initially, we adopt the same classification approach as in PN (Section~\ref{pn}) for the first few test batches to obtain a confidence score for each query sample, denoted as \( \text{Logit} \). For each class $c \in \{1, \dots, C\}$, given the normalized logits produced by the softmax function:
\begin{equation}
l_c = \frac{e^{\text{Logit}_c(\boldsymbol{x})}}{\sum_{j=1}^{C} e^{\text{Logit}_j(\boldsymbol{x})}}, \quad \text{where } \sum_{c=1}^{C} l_c = 1.
\end{equation}

To evaluate whether these samples should be added to the auxiliary set, we define a binary confidence score function \( \mathcal{G}(\boldsymbol{x}^q) \in \{0, 1\} \), which determines whether a query sample \( \boldsymbol{x}^q \) should be included. This decision is based on two orthogonal aspects of prediction confidence:

\paragraph{Global Confidence via Entropy.}

The first criterion captures the overall uncertainty of the model’s predictive distribution. The entropy of the predictive distribution is computed as:
\begin{equation}
\mathcal{H}(\boldsymbol{x}^q) = -\sum_{c=1}^{C} l_c \log l_c.
\end{equation}
A low entropy implies that the model assigns a high probability mass to a single class, indicating high confidence. We normalize the entropy to obtain a global confidence score:
\begin{equation}
\Delta = 1 - \frac{\mathcal{H}(\boldsymbol{x}^q)}{\log C},
\label{Delta}
\end{equation}
where \( \log C \) is the maximum entropy corresponding to a uniform distribution over \( C \) classes. Thus, \( \Delta \in [0, 1] \) measures the sharpness of the prediction, with higher values indicating greater certainty.

\paragraph{Local Confidence via Logit Margin.}

The second criterion evaluates the model’s ability to distinguish the predicted class from its nearest alternative. Let \( \hat{y} = \arg\max_c l_c \) denote the predicted class. We define the margin between the top-1 and top-2 probabilities as:
\begin{equation}
l_{\max} = \max_c l_c, \quad l_{\text{second}} = \max_{c \neq \hat{y}} l_c,
\end{equation}
\begin{equation}
\Delta' = \log\left(\frac{l_{\max}}{l_{\text{second}}}\right).
\label{Delta'}
\end{equation}
The normalized log-ratio \( \Delta' \in [0, 1] \) reflects the relative separation between the most likely class and its closest competitor, which serves as a local confidence indicator.

\paragraph{Sample Selection Rule.}

To ensure that only high-confidence samples contribute to prototype refinement, we include \( \boldsymbol{x}^q \) in the auxiliary set if and only if both confidence criteria are satisfied:
\begin{equation}
\mathcal{G}(\boldsymbol{x}^q) =
\begin{cases}
1 & \text{if } \Delta > \tau \text{ and } \Delta' > \tau', \\
0 & \text{otherwise},
\end{cases}
\label{criteria}
\end{equation}
where \( \tau \) and \( \tau' \) are predefined confidence thresholds. The auxiliary set is updated accordingly:
\begin{equation}
\text{if } \mathcal{G}(\boldsymbol{x}^q) = 1, \quad \mathcal{A}_{\hat{y}} \leftarrow \mathcal{A}_{\hat{y}} \cup \{\boldsymbol{x}^q\}; \quad \text{else: skip}.
\end{equation}

This dual-filtering mechanism ensures that only query samples with both globally confident and locally discriminative predictions are reused. It effectively reduces the risk of introducing noise into the prototype representation and improves the robustness of the few-shot learner.

\subsection{IPEC Testing Procedure}

After constructing an auxiliary set, we enrich class representations by storing confidently classified query samples. When a previously seen class (e.g., Class~1) reappears in batch \( n+1 \), we retrieve its auxiliary samples from \( \mathcal{A}_1 \) to refine the prototype. As illustrated in Fig.~\ref{fig:pipeline}, these samples provide additional representative features from past batches.

These auxiliary samples are incorporated into the prototype computation by averaging them with the current batch's support set. Specifically, given a support set \( \mathcal{S} = \bigcup_{c=1}^{C} \mathcal{S}_c \), where \( \mathcal{S}_c \) denotes support samples for class \( c \), the prototype \( \boldsymbol{p}^{aux}_c \in \mathbb{R}^d \) for class \( c \) is estimated by:

\begin{equation}
\boldsymbol{p}^{aux}_c = \frac{1}{|\mathcal{S}_c| + |\mathcal{A}_c|} \left( \sum_{(\boldsymbol{x}^{s}_i, y^{s}_i) \in \mathcal{S}_c} f(\boldsymbol{x}^{s}_i) + \sum_{\boldsymbol{x}^{a}_j \in \mathcal{A}_c} f(\boldsymbol{x}^{a}_j) \right).
\label{pc}
\end{equation}

This prototype \( \boldsymbol{p}^{aux}_c \) integrates both the current support samples and the previously stored auxiliary samples, making it a more representative estimation of the true class center. As a result, the refined prototype better aligns with the ground truth distribution of classes, leading to improved classification accuracy for query samples in the current batch.

Subsequently, for a query sample $\boldsymbol{x}^q$, the model computes the distance between its embedding and each class prototype, optionally computed by either Eq. (\ref{distance_euclid}) or Eq. (\ref{distance_cosine}). The predicted class is then:

\begin{equation}
\hat{y} = \arg\max_{c} \text{Logit}(\boldsymbol{x}^q,\boldsymbol{p}^{aux}_c).
\end{equation}

This formulation integrates both second-order class information and angular alignment, enabling more discriminative test-time inference.

By leveraging the auxiliary set, our approach adaptively enhances the prototype representation over time, mitigating the impact of limited support samples and improving classification performance in FSL scenarios. To clearly illustrate the overall workflow of our testing approach, we summarize the procedure in Algorithm~\ref{ipec}.

\begin{algorithm}
\caption{IPEC Testing Procedure with Auxiliary Set Updating}
\label{ipec}
\KwIn{Test batches $\{\mathcal{B}_n\}_{n=1}^N$, confidence thresholds $\tau$, $\tau'$, auxiliary sets $\{\mathcal{A}_c\}$ (initialized empty)}
\For{each test batch $\mathcal{B}_n$}{
    Extract support sets $\{\mathcal{S}_c\}_{c=1}^C$ and query sample $\boldsymbol{x}^q$ from the batch\;

    \For{each class $c$ in the batch}{
        Construct extended set $\mathcal{E}_c = \mathcal{S}_c \cup \mathcal{A}_c$\;

        Compute prototype $\boldsymbol{p}^{aux}_c$ from $\mathcal{E}_c$\;

    }

    \For{each class $c$}{
        Compute similarity logit $\text{Logit}_c(\boldsymbol{x}^q,\boldsymbol{p}^{aux}_c)$ using Euclidean distance or cosine similarity\;
    }

    Predict class $\hat{y} = \arg\max_c \text{Logit}_c(\boldsymbol{x}^q)$\;

    Compute softmax probabilities $l_c = \frac{e^{\text{Logit}_c}}{\sum_j e^{\text{Logit}_j}}$\;

    Compute entropy $\mathcal{H}(\boldsymbol{x}^q) = -\sum_c l_c \log l_c$\;

    Compute normalized global confidence $\Delta = 1 - \frac{\mathcal{H}(\boldsymbol{x}^q)}{\log C}$\;

    Identify $l_{\max} = \max_c l_c$, $l_{\text{second}} = \max_{c \neq \hat{y}} l_c$\;

    Compute log-ratio margin $\mathcal{D}_{\text{KL}} = \log\left(\frac{l_{\max}}{l_{\text{second}}}\right)$\;

    Compute normalized local confidence $\Delta' = \frac{\mathcal{D}_{\text{KL}}}{\log C}$\;

    \If{$\Delta > \tau$ \textbf{and} $\Delta' > \tau'$}{
        Add query sample to auxiliary set: $\mathcal{A}_{\hat{y}} \leftarrow \mathcal{A}_{\hat{y}} \cup \{\boldsymbol{x}^q\}$\;
    }
}
\KwOut{Predicted labels for all query samples}
\end{algorithm}

\subsection{Auxiliary Samples Removal Strategy}
\label{remove strategy}

To enhance the reliability of the auxiliary set, we designed and compared three update strategies. The most basic approach, referred to as ADD, follows the criterion described in Section~\ref{set construction}, where samples with logits above a predefined threshold are directly added to the auxiliary set of the predicted category. However, this naive method has two main limitations: (i) a sample may already exist in the auxiliary set, and (ii) the sample may have been previously misclassified into the wrong category. To address these issues, we propose two improved strategies REPLACE and REMOVE, which progressively refine the handling of potential “bad samples” to improve the quality of the auxiliary set and the overall model performance.

We compare three strategies for updating the auxiliary set:

\begin{itemize}
\item ADD: Simply adds the sample to the auxiliary set without verifying duplication or correctness.
\item REPLACE: Updates the stored category of a duplicated sample with the current predicted category, assuming later predictions are more reliable.
\item REMOVE: Removes duplicates samples that are previously misclassified to prevent potential propagation of noisy samples.
\end{itemize}

\subsection{Bayesian Interpretations of IPEC Design Components}

\subsubsection{IPEC Enables Continual Knowledge Accumulation}
\label{bayes}

From a Bayesian perspective, the auxiliary set \( \mathcal{A}_c \) can be interpreted as a growing posterior sample set that incrementally refines the estimation of the class-conditional feature mean. Specifically, the prototype for class \( c \) is estimated as computed in Eq. (\ref{pc}).
Assuming that samples are i.i.d. and that the confidence-based selection mechanism is reliable (i.e., governed by valid thresholds on \( \Delta \) and \( \Delta' \)), the prototype \( \boldsymbol{p}_c \) becomes a consistent estimator of the true class mean \( \mu_c = \mathbb{E}[f(\boldsymbol{x}) \mid y = c] \). According to the Law of Large Numbers \cite{murphy2012machine}:
\begin{equation}
 \boldsymbol{p}_c \xrightarrow{\text{a.s.}} \mu_c \quad \text{as} \quad |\mathcal{A}_c| \rightarrow \infty.   
 \label{lln}
\end{equation}

This indicates that even without updating network parameters, IPEC enables a form of continual adaptation via posterior refinement, gradually improving classification accuracy through the accumulation of high-confidence query samples.

\vspace{1em}

\subsubsection{Support Set as Prior Initialization}
\label{ini}

In the IPEC framework, the support set \( \mathcal{S}_c \) serves as the prior for estimating the initial class prototype. The prior-based initialization is given by:
\begin{equation}
    \boldsymbol{p}_c^{(0)} = \frac{1}{|\mathcal{S}_c|} \sum_{(\boldsymbol{x}^{s}_i, y^{s}_i) \in \mathcal{S}_c} f(\boldsymbol{x}^{s}_i).
\end{equation}

As test-time query samples are gradually filtered into \( \mathcal{A}_c \), according to Eq. (\ref{pc}), the prototype evolves:
\begin{equation}
    \text{Weight of } \mathcal{S}_c \searrow, \quad \text{Weight of } \mathcal{A}_c \nearrow.
\end{equation}

This progressive shift highlights the Bayesian nature of prototype refinement, where the influence of the prior (support set) fades as the posterior (auxiliary set) becomes dominant.

\vspace{1em}

\subsubsection{Warm-up Phase Analysis}

Although the Law of Large Numbers ensures Eq. (\ref{lln}), practical considerations impose limits on the size of \( \mathcal{A}_c(t) \). Hence, identifying the minimal auxiliary size that achieves a sufficiently accurate approximation of \( \mu_c \) becomes critical.

In the early phase of inference, the auxiliary set \( \mathcal{A}_c \) contains very few elements. At this stage, prototype estimates rely heavily on the support set \( \mathcal{S}_c \), which serves as the prior in the Bayesian interpretation. As query samples are gradually filtered into \( \mathcal{A}_c \), the prototype transitions from prior-dominated to posterior-dominated. This transition can be formalized by defining the prototype at step \( t \) as:

\begin{equation}
\boldsymbol{p}^{aux}_c(t) = \frac{1}{|\mathcal{S}_c| + |\mathcal{A}_c(t)|} \left( \sum_{(\boldsymbol{x}^{s}_i, y^{s}_i) \in \mathcal{S}_c} f(\boldsymbol{x}^{s}_i) + \sum_{\boldsymbol{x}^{a}_j \in \mathcal{A}_c(t)} f(\boldsymbol{x}^{a}_j) \right),
\end{equation}

\noindent where \( \mathcal{A}_c(t) \) denotes the auxiliary set constructed up to step \( t \). In the initial regime where \( |\mathcal{A}_c(t)| \ll |\mathcal{S}_c| \), the prototype can be approximated as:

\begin{equation}
    \boldsymbol{p}_c(t) \approx \boldsymbol{p}_c^{(0)} + \frac{|\mathcal{A}_c(t)|}{|\mathcal{S}_c|} \left( \boldsymbol{\mu}_c^{\text{aux}}(t) - \boldsymbol{p}_c^{(0)} \right),
\end{equation}

\noindent where \( \boldsymbol{\mu}_c^{\text{aux}}(t) = \frac{1}{|\mathcal{A}_c(t)|} \sum_{\boldsymbol{x}^{a}_j \in \mathcal{A}_c(t)} f(\boldsymbol{x}^{a}_j) \) is the mean embedding of the auxiliary set. Although this expression highlights the increasing influence of \( \mathcal{A}_c(t) \) over time, it ignores the fact that auxiliary samples are selectively incorporated based on confidence thresholds. Selected via confidence thresholds \( \Delta \) and \( \Delta' \), auxiliary samples are more likely near the true class center \( \mu_c \). This makes \( \mathcal{A}_c(t) \) a more informative, posterior-informed representation than randomly chosen support samples.

To understand prototype quality, we still consider its deviation from the true class mean:

\begin{align}
    \|\boldsymbol{p}_c(t) - \mu_c\|^2 
    &\approx
    \|\boldsymbol{p}_c^{(0)} - \mu_c\|^2 \notag \\
    &\quad + \mathbb{E}[\| \boldsymbol{\mu}_c^{\text{aux}}(t) - \mu_c \|^2] \cdot \left( \frac{|\mathcal{A}_c(t)|}{|\mathcal{S}_c|} \right)^2 \notag \\
    &\quad + 2 \cdot \text{Cov}\left( \boldsymbol{p}_c^{(0)} - \mu_c, \boldsymbol{\mu}_c^{\text{aux}}(t) - \mu_c \right),
\end{align}

\noindent where the second term reflects the posterior variance and the third term is the error covariance. Since the auxiliary set accumulates only high-confidence examples, we assume the posterior variance term decays rapidly with \( t \). This allows a stable estimate to be formed while \( |\mathcal{A}_c(t)| \) remains small, aligning with our goal of finding a minimal sufficient set.

Based on the above analysis, there exists a warm-up period, which we can understand as two distinct phases separated by a critical threshold \( w \). The first phase, the Posterior Concentration Stage (\( t < w \)), is the initial period where the auxiliary set \( \mathcal{A}_c(t) \) is built until it becomes sufficiently informative to support accurate and stable prototype estimation. During this phase, the estimation shifts from being prior-dominant to posterior-dominant. 
Formally, this threshold \( w \) represents the minimal number of steps at which the auxiliary set's quality significantly surpasses that of the prior, as defined by the condition:
\begin{equation}
\mathbb{E}\left[\left\|\boldsymbol{\mu}_c^{\text{aux}}(w) - \mu_c\right\|^2\right] \ll \left\|\boldsymbol{p}_c^{(0)} - \mu_c\right\|^2.
\end{equation}

That is, at step \( w \), the auxiliary set has become sufficiently mature that its empirical mean \( \boldsymbol{\mu}_c^{\text{aux}}(w) \) provides a much closer approximation to the true class mean \( \mu_c \) than the prior-based prototype \( \boldsymbol{p}_c^{(0)} \). This marks the onset of posterior-driven estimation in the IPEC framework.

The second phase, the Saturation Stage (\( t \geq w \)), begins once this threshold is crossed. As more confident query samples are incorporated, the auxiliary set approaches a semantically complete representation of the class. Beyond this point, further expansion yields diminishing improvements in accuracy, and the prototype estimate simplifies to:

\begin{equation}
    \boldsymbol{p}_c(t) \approx \boldsymbol{\mu}_c^{\text{aux}}(t) + \epsilon,
\end{equation}

\noindent where \( \epsilon \) is a diminishing correction term scaled by \( |\mathcal{S}_c| / |\mathcal{A}_c(t)| \). 

These two sub-stages jointly constitute the entire warm-up process. Importantly, once the warm-up period \( w \) concludes, the auxiliary set \( \mathcal{A}_c(w) \) can be regarded as the minimal auxiliary set that sufficiently approximates the true class mean \( \mu_c \), echoing the core goal stated at the beginning of this section.

Inspired by the above theoretical framework, we propose a practical two-stage inference protocol:

\begin{itemize}
    \item Warm-up Stage: The model focuses exclusively on constructing the auxiliary set \( \mathcal{A}_c \) by selectively accepting high-confidence query samples. During this phase, no accuracy is recorded, and prototype updates are driven solely by posterior accumulation.
    
    \item Testing Stage: Once the auxiliary set becomes sufficiently mature and stable, it is fixed. The model switches to evaluation mode, where no further updates are made to the prototype, and accuracy is recorded based on the fixed posterior-driven representation.
\end{itemize}

This two-stage design not only aligns with the theoretical conditions for accurate posterior estimation, but also addresses the practical constraint on the size of the auxiliary set, ensuring both computational efficiency and semantic robustness in real-world deployment scenarios.

\section{Experiments}
\label{sec:experiments}
\subsection{Datasets}

Our work conducts experiments on four common FSL benchmarks, MiniImageNet \cite{Vinyals2016}, TieredImageNet \cite{Ren2018}, CIFAR \cite{krizhevsky2009learning} and FC100 \cite{oreshkin2018tadam}, in which MiniImageNet and TieredImageNet are derivatives of the ImageNet-1K dataset \cite{Russakovsky2015}. And we also conduct cross-domain FSL tests on datasets ChestX \cite{wang2017chestx}, ISIC \cite{tschandl2018ham10000}, EuroSAT \cite{helber2019eurosat}, CropDiseases \cite{mohanty2016using}, CUB \cite{Wah2011}, Stanford Cars \cite{krause20133d}, Places \cite{zhou2017places} and Plantae \cite{van2018inaturalist}.

\begin{table}[h]
\centering
\caption{Dataset information. In-domain datasets have class splits in the format train:val:test and are used for training and evaluation. Cross-domain datasets have a single number in the Classes column and are used for testing only}
\label{tab:dataset_info}
\begin{tabular}{@{}cccccl@{}}
\toprule
Dataset & Domain & Classes & Images \\
\midrule
MiniImageNet & General recognition & 64:16:20 & 60,000 \\
TieredImageNet & General recognition & 351:97:160 & 779,165 \\
CIFAR & General recognition & 64:16:20 & 60,000 \\
FC100 & General recognition & 64:16:20 & 60,000 \\
\midrule
CUB & Fine - grained bird recognition & 50 & 2,953 \\
Cars & Fine - grained car recognition & 49 & 2,027 \\
Plantae & Plantae recognition & 50 & 17,253 \\
Places & Scene recognition & 19 & 3,800 \\
CropDiseases & Agricultural disease recognition & 38 & 43,456 \\
EuroSAT & Satellite imagery recognition & 10 & 27,000 \\
ISIC & Skin - lesion recognition & 7 & 10,015 \\
ChestX & X - ray chest recognition & 7 & 25,847 \\
\bottomrule
\end{tabular}
\end{table}

\subsection{Setting}
\label{sec:5.2}

Our experimental setup is identical to that of SemFew \cite{zhang2024simple}. We use ResNet12 and Swin-T as backbones. Specifically, for ResNet12, the input size is 32×32 for CIFAR and FC100, while it is 84×84 for all other datasets. For Swin-T, the input size is 224×224 for all datasets.

We evaluate our proposed modification to the PN testing process under 5-way 1-shot and 5-shot settings using two distinct configurations, denoted as \textbf{Ours} and \textbf{Ours*}. Specifically, for consistency with prior work and to establish a direct comparison, the \textbf{Ours} setting deliberately omits the warm-up phase and directly evaluates performance on 600 randomly sampled tasks from the novel set. In contrast, the \textbf{Ours}* setting fully embodies our proposed two-stage protocol: it first performs a warm-up phase to construct a robust auxiliary set, after which this set is frozen for a subsequent, more stable test phase. For both configurations, we report the mean accuracy and 95\% confidence interval over the 600 test tasks, each containing 15 query samples per class.

\subsection{Baseline}

SOTA methods are applied to the few-shot classification task to facilitate a thorough comparison, including CME \cite{zhou2023learning}, MetaDiff \cite{zhang2024metadiff}, SEGA+AFR \cite{zhu2024boosting}, SEVPro \cite{Cai2024}, ALFA+MeTAL \cite{baik2023learning}, SemFew \cite{zhang2024simple}, PBML \cite{fu2024prototype}, CDFT \cite{zhang2024cross}, as well as cross-domain SOTA methods like Dara \cite{zhao2023dual}, FSC \cite{ji2025frequency}, SVasP \cite{li2025svasp}, fine-tune \cite{guo2020broader}, LDPNet \cite{zhou2023revisiting}, FLoR \cite{zou2024flatten}, StyleAdv-FT \cite{fu2023styleadv}, FSViT \cite{song2024matching}.


\begin{table*}[!t]
\caption{Comparison of our method with SOTA approaches on four benchmark datasets: MiniImageNet, TieredImageNet, CIFAR-FS, and FC100.}
\label{tab:main_results}
\centering
\footnotesize 
\setlength{\tabcolsep}{3.0pt} 
\begin{tabular}{l|l|cc|cc|cc|cc}
\toprule
\multirow{2}{*}{\textbf{Models}} & \multirow{2}{*}{\textbf{Backbone}} & \multicolumn{2}{c|}{\textbf{MiniImageNet}} & \multicolumn{2}{c|}{\textbf{TieredImageNet}} & \multicolumn{2}{c|}{\textbf{CIFAR-FS}} & \multicolumn{2}{c}{\textbf{FC100}} \\ 
\cmidrule(lr){3-4} \cmidrule(lr){5-6} \cmidrule(lr){7-8} \cmidrule(lr){9-10}
 &  & 1-shot & 5-shot & 1-shot & 5-shot & 1-shot & 5-shot & 1-shot & 5-shot \\

\midrule
CME (TCSVT’23) \cite{zhou2023learning} & ResNet-12 & 63.01$\pm$0.20 & 79.78$\pm$0.14 & 67.18$\pm$0.23 & 82.44$\pm$0.16 & 72.63$\pm$0.31 & 85.88$\pm$0.15 & 42.05$\pm$0.35 & 57.56$\pm$0.17 \\
MetaDiff (AAAI’24) \cite{zhang2024metadiff} & ResNet-12 & 64.99$\pm$0.77 & 81.21$\pm$0.56 & 72.33$\pm$0.92 & 86.31$\pm$0.62 & - & - & - & - \\
SEGA+AFR (AAAI’24) \cite{zhu2024boosting} & ResNet-12 & 71.14$\pm$0.60 & 84.26$\pm$0.42 & 72.87$\pm$0.45 & 85.26$\pm$0.54 & - & - & - & - \\
SEVPro (IJCAI’24) \cite{Cai2024} & ResNet-12 & 71.81$\pm$0.22 & 78.88$\pm$0.18 & 72.77$\pm$0.30 & 84.04$\pm$0.21 & 80.36$\pm$0.24 & 86.12$\pm$0.20 & - & - \\
ALFA+MeTAL (TPAMI’24) \cite{baik2023learning} & ResNet-12 & 66.61$\pm$0.28 & 81.43$\pm$0.25 & 70.29$\pm$0.40 & 86.17$\pm$0.35 & 76.32$\pm$0.43 & 86.73$\pm$0.31 & 44.54$\pm$0.50 & 58.44$\pm$0.42 \\
SemFew (CVPR’24) \cite{zhang2024simple} & ResNet-12 & 77.63$\pm$0.63 & 83.04$\pm$0.48 & 78.96$\pm$0.80 & 85.88$\pm$0.62 & 83.65$\pm$0.70 & 87.66$\pm$0.60 & 54.36$\pm$0.71 & 62.79$\pm$0.74 \\
PBML (TNNLS’25) \cite{fu2024prototype} & ResNet-12 & 63.60$\pm$0.70 & 81.94$\pm$0.44 & 70.64$\pm$0.72 & 85.39$\pm$0.40 & 73.07$\pm$0.59 & 85.51$\pm$0.41 & 47.92$\pm$0.49 & 62.96$\pm$0.51 \\
CDFT (TCSVT’25) \cite{zhang2024cross} & ResNet-12 & 68.18$\pm$\text{N/A} & 83.74$\pm$\text{N/A} & 73.20$\pm$\text{N/A} & 86.91$\pm$\text{N/A} & 75.03$\pm$\text{N/A} & 86.78$\pm$\text{N/A} & 42.46$\pm$\text{N/A} & 57.44$\pm$\text{N/A} \\
\midrule
Baseline & ResNet-12 & 61.58$\pm$0.86 & 77.90$\pm$0.64 & 63.65$\pm$0.92 & 79.87$\pm$0.69 & 71.59$\pm$0.88 & 85.94$\pm$0.62 & 42.97$\pm$0.74 & 58.92$\pm$0.78 \\
\rowcolor{lightblue} Ours & ResNet-12 & 83.98$\pm$0.52 & 84.04$\pm$0.52 & 81.88$\pm$0.90 & 85.68$\pm$0.56 & 90.87$\pm$0.45 & 91.06$\pm$0.42 & 71.31$\pm$0.73 & 72.06$\pm$0.66 \\
\rowcolor{lightblue} Ours* & ResNet-12 & \textbf{84.79$\pm$0.50} & \textbf{84.79$\pm$0.50} & \textbf{86.52$\pm$0.48} & \textbf{86.98$\pm$0.49} & \textbf{91.26$\pm$0.43} & \textbf{91.21$\pm$0.43} & \textbf{73.41$\pm$0.61} & \textbf{73.09$\pm$0.60} \\

\midrule
SemFew (CVPR’24) \cite{zhang2024simple} & Swin-T & 78.94$\pm$0.66 & 86.49$\pm$0.50 & 82.37$\pm$0.77 & 89.89$\pm$0.52 & 84.34$\pm$0.67 & 89.11$\pm$0.54 & 54.27$\pm$0.77 & 65.02$\pm$0.72 \\
\midrule
Baseline & Swin-T & 62.73$\pm$0.54 & 78.11$\pm$0.63 & 69.53$\pm$0.90 & 84.73$\pm$0.61 & 76.87$\pm$0.85 & 88.50$\pm$0.59 & 45.41$\pm$0.70 & 63.19$\pm$0.76 \\
\rowcolor{lightblue} Ours & Swin-T & 83.81$\pm$0.49 & 84.14$\pm$0.47 & 86.86$\pm$0.81 & 89.96$\pm$0.47 & 92.45$\pm$0.44 & 92.69$\pm$0.40 & 76.43$\pm$0.70 & 77.01$\pm$0.64 \\
\rowcolor{lightblue} Ours* & Swin-T & \textbf{84.21$\pm$0.47} & \textbf{84.23$\pm$0.47} & \textbf{90.39$\pm$0.42} & \textbf{90.58$\pm$0.42} & \textbf{92.61$\pm$0.38} & \textbf{92.68$\pm$0.38} & \textbf{78.55$\pm$0.59} & \textbf{78.59$\pm$0.58} \\
\bottomrule
\end{tabular}
\end{table*}

\begin{table*}[ht]
\caption{Comparison of our method with previous approaches under cross-domain setting}
\label{tab:CUB}
\centering
\footnotesize 
\setlength{\tabcolsep}{7.7pt} 
\begin{tabular}{lcccccccccc}
\toprule
models          & Backbone   & ChestX & ISIC   & EuroSAT & CropDiseases & CUB   & Cars   & Places & Plantae & AVG       \\
\midrule
Dara(TPAMI23) \cite{zhao2023dual}         & ResNet-10 & 22.93  & 38.49  & 69.39   & 81.50       & 52.70 & 35.25  & 51.25  & 42.08   & 49.20     \\
FSC(TIP25) \cite{ji2025frequency}         & ResNet-10 & 23.29  & 35.22  & 69.77   & 77.30       & 55.26 & 36.34  & 58.95  & 41.33   & 49.68     \\
SVasP(AAAI25) \cite{li2025svasp}          & ResNet-10 & 23.23  & 37.63  & 72.30   & 77.45       & 49.49 & 38.18  & 59.07  & 41.22   & 49.82     \\
\midrule
fine-tune(ECCV2020) \cite{guo2020broader} & ResNet-10 & 22.50  & 32.92  & 62.87   & 68.81       & 42.18 & 33.35  & 49.15  & 39.17   & 43.87     \\
\rowcolor{lightblue} fine-tune + Ours     & ResNet-10 & 22.67  & 43.32  & 66.61   & 86.23       & 74.25 & 69.14  & 62.41  & 64.75   & 61.17     \\
LDPNet(CVPR24) \cite{zhou2023revisiting}  & ResNet-10 & 23.02  & 33.96  & 64.55   & 69.15       & 46.74 & 33.86  & 53.08  & 41.14   & 45.69     \\
\rowcolor{lightblue} LDPNet + Ours        & ResNet-10 & 24.22  & 45.77  & 84.45   & 93.92       & 83.62 & 67.16  & 75.37  & 74.57   & 68.64     \\
FLoR(AAAI25) \cite{zou2024flatten}        & ResNet-10 & 22.74  & 36.39  & 63.18   & 69.88       & 49.23 & 32.50  & 50.10  & 41.65   & 45.71     \\
\rowcolor{lightblue} FLoR + Ours          & ResNet-10 & 23.13  & 42.31  & 66.16   & 82.56       & 59.82 & 40.29  & 60.44  & 49.90   & 53.08     \\
\midrule
Baseline                                  & ViT     & 22.28  & 29.88  & 61.58   & 68.51       & 49.84 & 33.38  & 58.29  & 43.96   & 45.97     \\
StyleAdv-FT(CVPR23) \cite{fu2023styleadv} & ViT     & 22.92  & 33.99  & 74.93   & 84.11       & 84.01 & 40.48  & 72.64  & 55.52   & 58.58     \\
FSViT(TCSVT24) \cite{song2024matching}    & ViT     & 22.95  & 34.83  & 74.95   & 84.21       & -     & -      & -      & -       & 54.24     \\
\rowcolor{lightblue} Ours                 & ViT     & 25.85  & 34.37  & 83.42   & 90.11       & 83.39 & 67.98  & 78.11  & 64.70   & 65.99     \\
\bottomrule
\end{tabular}
\end{table*}

\begin{table*}[!h]
\caption{Ablation study of updating the auxiliary set with Swin-T}
\label{table:update}
\centering
\small
\begin{tabular}{
>{\centering\arraybackslash}p{1.23cm}| 
>{\centering\arraybackslash}p{1.42cm}|
>{\centering\arraybackslash}p{1.42cm}|
>{\centering\arraybackslash}p{1.42cm}|
>{\centering\arraybackslash}p{1.60cm}|
>{\centering\arraybackslash}p{1.42cm}|
>{\centering\arraybackslash}p{1.42cm}|
>{\centering\arraybackslash}p{1.42cm}|
>{\centering\arraybackslash}p{1.42cm}|
>{\centering\arraybackslash}p{0.8cm}
}
\toprule
\multirow{2}{*}{Models} & \multicolumn{2}{c|}{MiniImageNet} & \multicolumn{2}{c|}{TieredImageNet} & \multicolumn{2}{c|}{CIFAR} & \multicolumn{2}{c|}{FC100} & \multirow{2}{*}{AVG} \\ 
\cmidrule{2-3}\cmidrule{4-5}\cmidrule{6-7}\cmidrule{8-9}
 & 1-shot & 5-shot & 1-shot & 5-shot & 1-shot & 5-shot & 1-shot & 5-shot &  \\
\midrule
ADD      & 84.18 & 84.06 & 90.34 & 90.53 & 92.44 & 92.54 & 78.24 & 78.42 & 86.34 \\
REPLACE  & 84.20 & 84.08 & 90.38 & 90.57 & 92.49 & 92.64 & 78.16 & 78.29 & 86.35 \\
\rowcolor{lightblue} REMOVE   & \textbf{84.21} & \textbf{84.23} & \textbf{90.39} & \textbf{90.58} & \textbf{92.61} & \textbf{92.68} & \textbf{78.55} & \textbf{78.59} & \textbf{86.48} \\
\midrule
\midrule
Models & ChestX & ISIC & EuroSAT & CropDiseases & CUB & Cars & Places & Plantae & AVG \\ 
\midrule
ADD & 22.86 & 34.58 & 83.54 & 90.03 & 73.70 & \textbf{44.22} & 78.11 & 64.70 & 61.47 \\
REPLACE & 22.86 & 35.04 & 83.56 & 90.04 & 73.63 & 43.22 & 79.10 & 65.02 & 61.56 \\
\rowcolor{lightblue} REMOVE & 22.86 & \textbf{35.34} & \textbf{83.59} & \textbf{90.05} & \textbf{73.56} & 42.28 & \textbf{79.38} & \textbf{65.69} & \textbf{61.59} \\
\bottomrule
\end{tabular}
\end{table*}

\begin{table}
    \caption{Ablation study of confidence-based filtering mechanism}
    \label{tab:re1}
    \centering
    \fontsize{9}{11}\selectfont
    \begin{tabular}{c c ccll}
    \toprule
    Local & Global &  Mini & Tiered & CIFAR & FC100 \\
    \midrule 
    \ding{55} & \ding{55} & 84.28 & 86.48 & 91.13 & 73.35 \\ 
    \ding{55} & \ding{51} & 84.35 & 86.49 & 91.13 & 73.37 \\ 
    \ding{51} & \ding{55} & 84.71 & 86.50 & 91.26 & 73.36 \\ 
    \rowcolor{lightblue} \ding{51} & \ding{51} & \textbf{84.79} & \textbf{86.52} & \textbf{91.26}& \textbf{73.41}  \\
    \bottomrule  
    \end{tabular}
\end{table}

\subsection{Overall Experimental Validation}

\subsubsection{In-domain Comparison with SOTA Methods}

As depicted in Table \ref{tab:main_results}, our proposed methodology exhibits remarkable superiority over existing SOTA approaches across diverse datasets and backbone architectures.

On both MiniImageNet and TieredImageNet, our method with a ResNet-12 backbone yields outstanding results. For instance, in the MiniImageNet 1-shot setting, our method achieves an accuracy of $83.98\% \pm 0.52$, significantly outperforming prior works such as SemFew ($77.63\% \pm 0.63$). The \textbf{Ours*} setting further pushes the SOTA to $84.79\% \pm 0.50$ on MiniImageNet and $86.52\% \pm 0.48$ on TieredImageNet in the 1-shot setting. When employing the Swin-T backbone, our method consistently surpasses competitive baselines and sets new SOTA records. Specifically, on TieredImageNet, the \textbf{Ours*} variant achieves remarkable accuracies of $90.39\% \pm 0.42$ (1-shot) and $90.58\% \pm 0.42$ (5-shot). Similarly, on MiniImageNet, it obtains a top-tier 1-shot accuracy of $84.21\% \pm 0.47$, further demonstrating the robustness and generalization capabilities of our approach.

Likewise, our method also excels on CIFAR and FC100, establishing new SOTA results. With a ResNet-12 backbone, the \textbf{Ours*} setting achieves $91.26\% \pm 0.43$ on CIFAR 1-shot and $73.41\% \pm 0.61$ on FC100 1-shot, demonstrating a substantial lead over all competitors. When utilizing the Swin-T backbone, our method further solidifies its superiority, reaching top accuracies of $92.61\% \pm 0.38$ on CIFAR and a commanding $78.55\% \pm 0.59$ on FC100 in the 1-shot setting.

In summary, these results unequivocally demonstrate that our method, irrespective of the backbone architecture, consistently outperforms existing methods in in-domain scenarios, thereby validating the effectiveness of our proposed approach.

\vspace{1em}

\subsubsection{Cross-domain Comparison with SOTA Methods}
In the context of cross-domain applications, Table~\ref{tab:CUB} provides a comprehensive comparison between our proposed method and several SOTA approaches across diverse datasets.

Compared with traditional methods such as Dara~\cite{zhao2023dual}, FSC~\cite{ji2025frequency}, and SVasP~\cite{li2025svasp}, all of which employ the ResNet-10 backbone, our method achieves significant performance improvements. When integrated with existing models, our approach further enhances their effectiveness, and the role of the backbone becomes more nuanced. For instance, the baseline fine-tune method~\cite{guo2020broader}, which adopts ResNet-10, yields an average accuracy of 43.87\%; after incorporating our method, the accuracy increases to 61.17\% while maintaining the same backbone. Similarly, for LDPNet~\cite{zhou2023revisiting}, also based on ResNet-10, our integration leads to a notable increase in average accuracy from 45.69\% to 68.64\%. Even for methods with relatively strong performance, such as StyleAdv-FT~\cite{fu2023styleadv} which leverages a ViT backbone and achieves an average accuracy of 58.58\%, our method using ViT surpasses it with an accuracy of 65.99\%. These results collectively demonstrate the robustness and adaptability of IPEC across different backbone configurations.
\begin{figure}[!t]
  \centering
  \vspace{-1mm} 
  \includegraphics[width=1.0\linewidth]{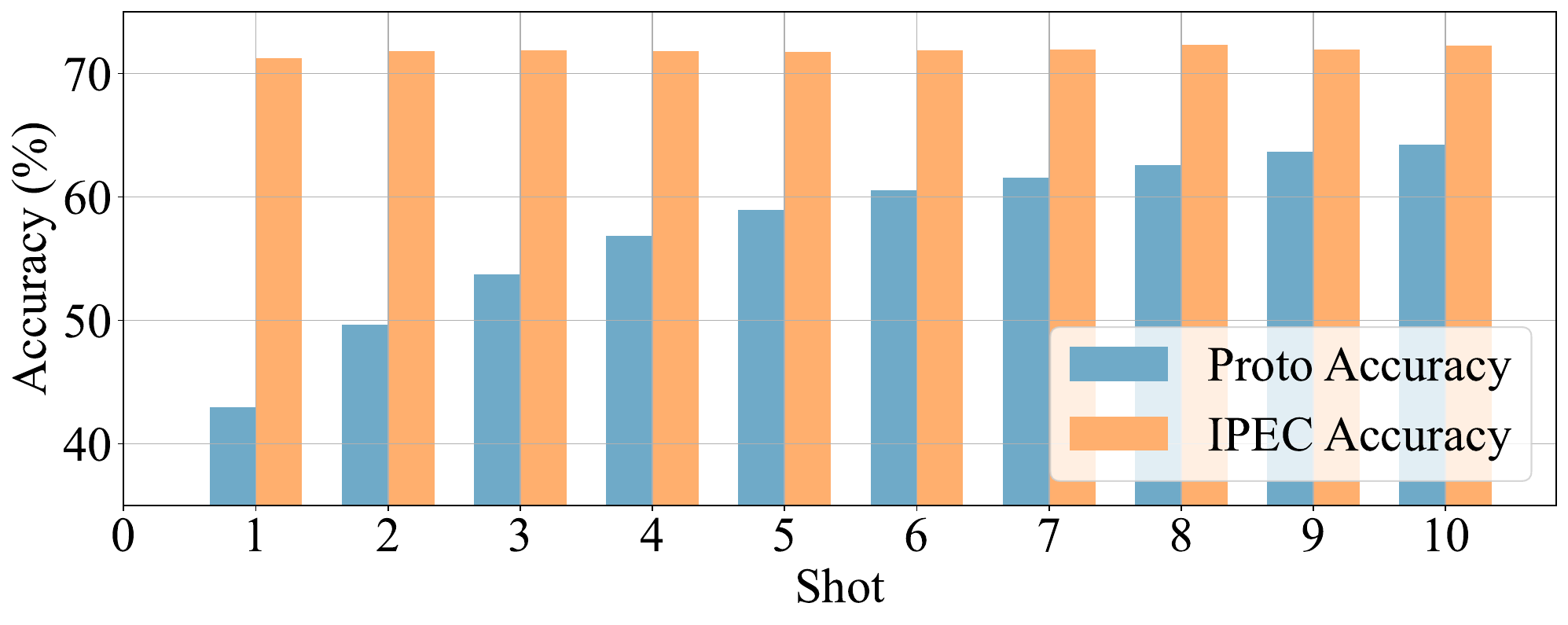} %
  \vspace{-5mm}  
  \caption{ Impact of the Number of Shots on Accuracy. IPEC demonstrates robust performance that is less sensitive to the number of shots.}
  \label{fig:shot}
  \vspace{-3mm}  
\end{figure}

\vspace{1em}

\subsubsection{Ablation Studies}

\paragraph{Updating Strategy of the Auxiliary Set} 

To enhance the reliability of the auxiliary set and consequently improve the performance of our model. The results of the ablation experiments, summarized in Table~\ref{table:update}, provide valuable insights into the effectiveness of these strategies.

Notably, the REPLACE strategy, which updates the category labels of duplicated samples based on predictions from later batches, outperforms the ADD strategy. This indicates that leveraging the refined knowledge accumulated in successive batches to correct potential misclassifications in the auxiliary set can effectively enhance model performance. More importantly, the REMOVE strategy, which completely discards misclassified and duplicated samples, consistently achieves the highest accuracy among the three.

These findings highlight two key observations: first, correcting misclassified samples with later predictions (REPLACE) outperforms the naive addition approach (ADD); second, and more critically, completely removing misclassified samples (REMOVE) leads to the best accuracy. This suggests that discarding potentially noisy or ambiguous samples is more beneficial than attempting to correct their labels. As demonstrated in Section~\ref{bayes}, classification accuracy generally improves over successive batches due to knowledge accumulation. Therefore, strategies that dynamically correct or remove erroneous samples, such as REMOVE, can effectively leverage this improvement to refine the auxiliary set, thereby enhancing its overall quality and ultimately boosting the model's performance.

\paragraph{Confidence-based Filtering Mechanism} 

To validate the effectiveness of the dual-confidence filtering mechanism in our IPEC framework, we conduct an ablation study focusing on the global confidence metric (entropy-based) and the local confidence metric (logit-margin-based). The results across four standard benchmarks are summarized in Table~\ref{tab:re1}.

The ablation study confirms that jointly using local and global confidence is crucial for our method's success. Disabling both metrics causes a performance drop, while the full model consistently achieves the best results, demonstrating their complementary roles. Essentially, global confidence selects high-quality samples, and local confidence resolves inter-class confusion, leading to more robust prototype refinement.

\begin{figure*}[!t]
\centering
\subfloat[]{\includegraphics[width=0.49\linewidth]{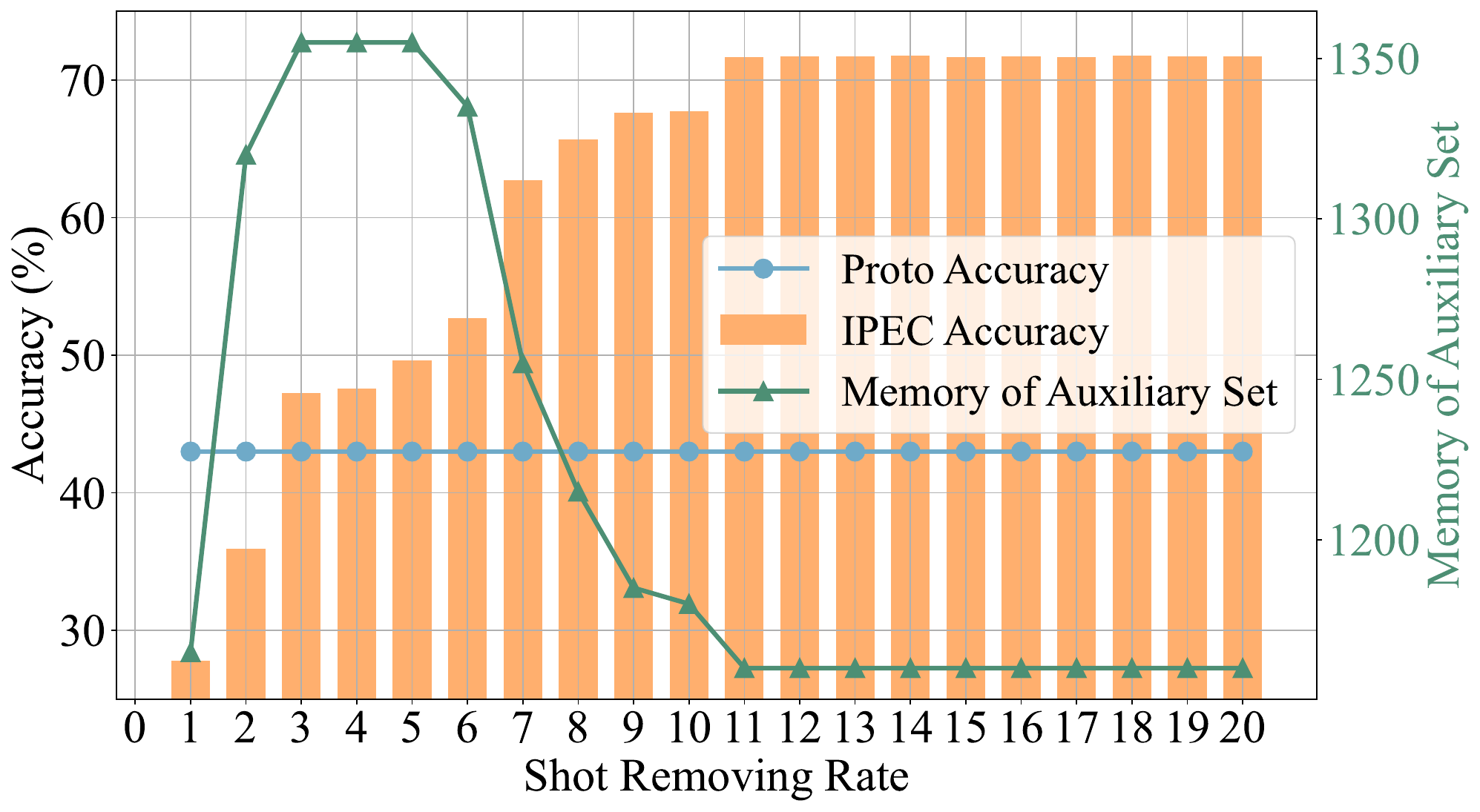}%
\label{fig:remove1}}
\hfil 
\subfloat[]{\includegraphics[width=0.49\linewidth]{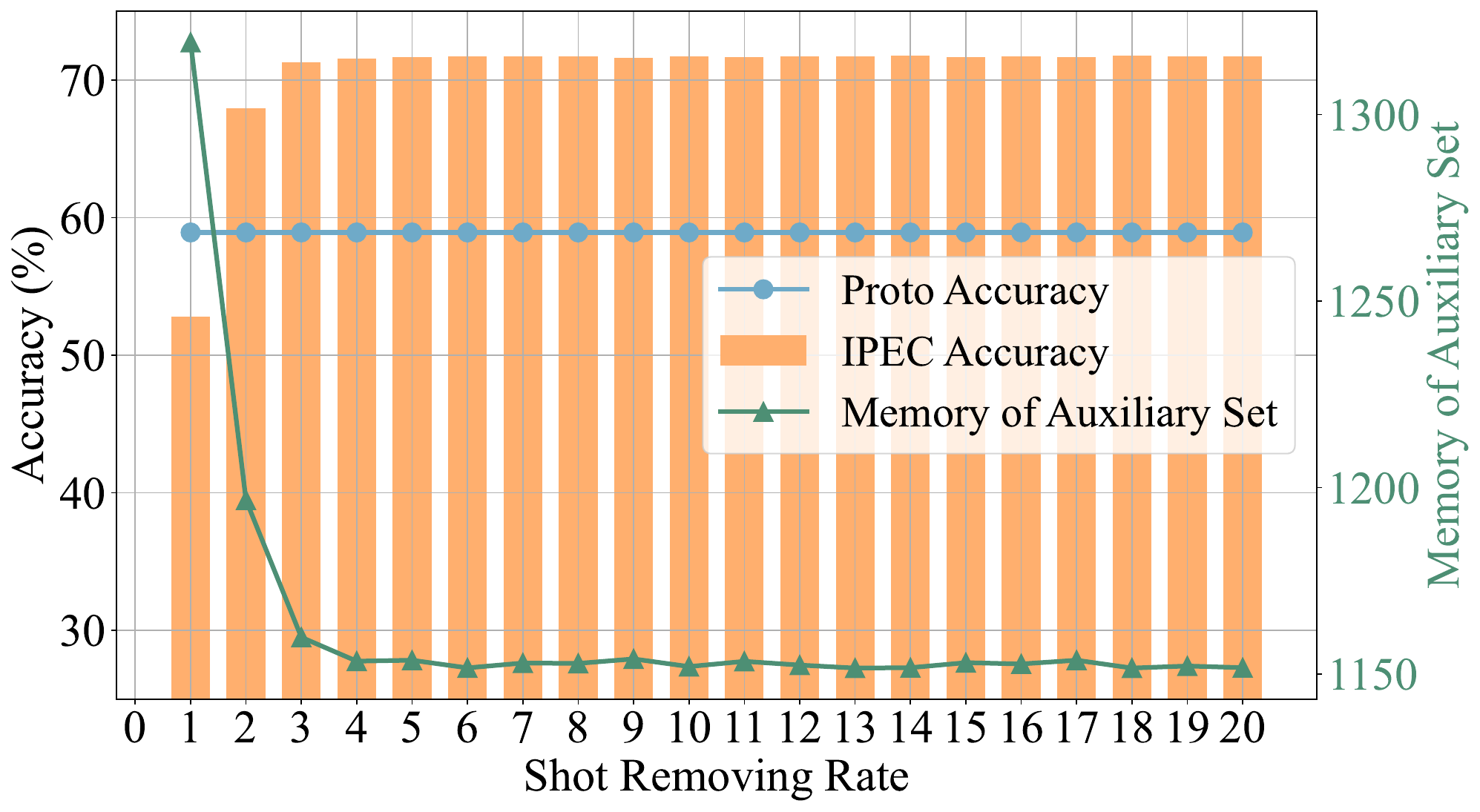}%
\label{fig:remove5}}
\caption{Analysis of IPEC's Dynamic Reliance on Support Shots. The figure illustrates how IPEC's accuracy and the auxiliary set's memory consumption evolve as support shots are gradually removed at different rates (one shot removed every k batches). (a) 1-shot setting (b) 5-shot setting}
\label{fig:remove15}
\end{figure*}

\vspace{1em}

\subsection{Empirical Support for Bayesian Interpretations of IPEC Design}

\subsubsection{Impact of Shots in IPEC}

As shown in Fig.~\ref{fig:shot}, PN benefits significantly from the increase in shots, as more support samples provide richer class information to the model, enabling more reliable and accurate classification. In contrast, the improvement observed in IPEC is relatively marginal.

The reason behind this phenomenon, as discussed in Section~\ref{ini}, is that in the IPEC method, the influence of support shots is mainly pronounced during the early batches but becomes negligible in later stages. Essentially, the support shots act as an initialization for the model training process. This explains why, as observed in Table~\ref{tab:main_results}, the accuracy gap between 5-shot and 1-shot settings in IPEC is considerably smaller than that of other existing methods.

To validate this, we conducted experiments illustrated in Fig.~\ref{fig:remove15}. The x-axis denotes the rate of support shot removal (one shot is removed every $k$ batches until the number of support shots reaches zero) meaning that prototypes are ultimately generated entirely from auxiliary samples. The initial number of shots is set to 1 and 5 in Fig.~\ref{fig:remove1} and Fig.~\ref{fig:remove5}, respectively.

The blue line represents the accuracy of PN under the same conditions, serving as a reference baseline. A consistent trend is observed in both figures: IPEC's accuracy is lower with high shot removal rates but rapidly increases and stabilizes as the removal frequency decreases. This indicates that in the early training stage, having more support shots helps better initialize the auxiliary set, which subsequently enhances accuracy in later batches. As $k$ increases, the slope of the accuracy curve gradually flattens, implying that once the auxiliary set matures, the impact of support shots significantly diminishes. At this point, removing support shots has little to no effect on final accuracy, further confirming that IPEC progressively shifts its reliance from support shots to auxiliary samples during the learning process. A comparison of the two figures reveals that IPEC exhibits greater stability and less sensitivity to support sample removal in the 5-shot setting than in the 1-shot setting. This is because the richer initial information in the 5-shot scenario facilitates faster maturation of the auxiliary set, thereby reducing the model's early-stage dependency on the support set and leading to more robust overall performance.

Moreover, the green dashed line in the figure (Memory) reflects quality and efficiency of the auxiliary set, showing distinct trends. In the 1-shot scenario, Memory first increases and then decreases. Initially, premature support removal leads to a small, noisy set due to low model confidence. As confidence grows, the set expands (Memory increases). After peaking, the now-proficient model creates a more efficient and compact representation, reducing the set's size and redundancy (Memory decreases). In contrast, the 5-shot scenario starts with a large, representative set, causing Memory to consistently decrease as it is progressively refined for higher efficiency. This demonstrates that: (i) premature support removal degrades auxiliary set quality, and (ii) richer initial support is crucial for building a mature, low-redundancy set, which enhances model robustness and generalization.

\vspace{1em}

\subsubsection{Warm-up Phase Effectiveness}

To validate the effectiveness of our proposed two-stage inference protocol, we conduct a series of experiments focused on the warm-up phase. This phase aims to construct a reliable auxiliary set from early incoming data before transitioning to a fixed-prototype inference stage.

\paragraph{How the duration of the warm-up phase affects model performance and memory usage}

\begin{figure}[t]
  \centering
  \includegraphics[width=0.95\linewidth]{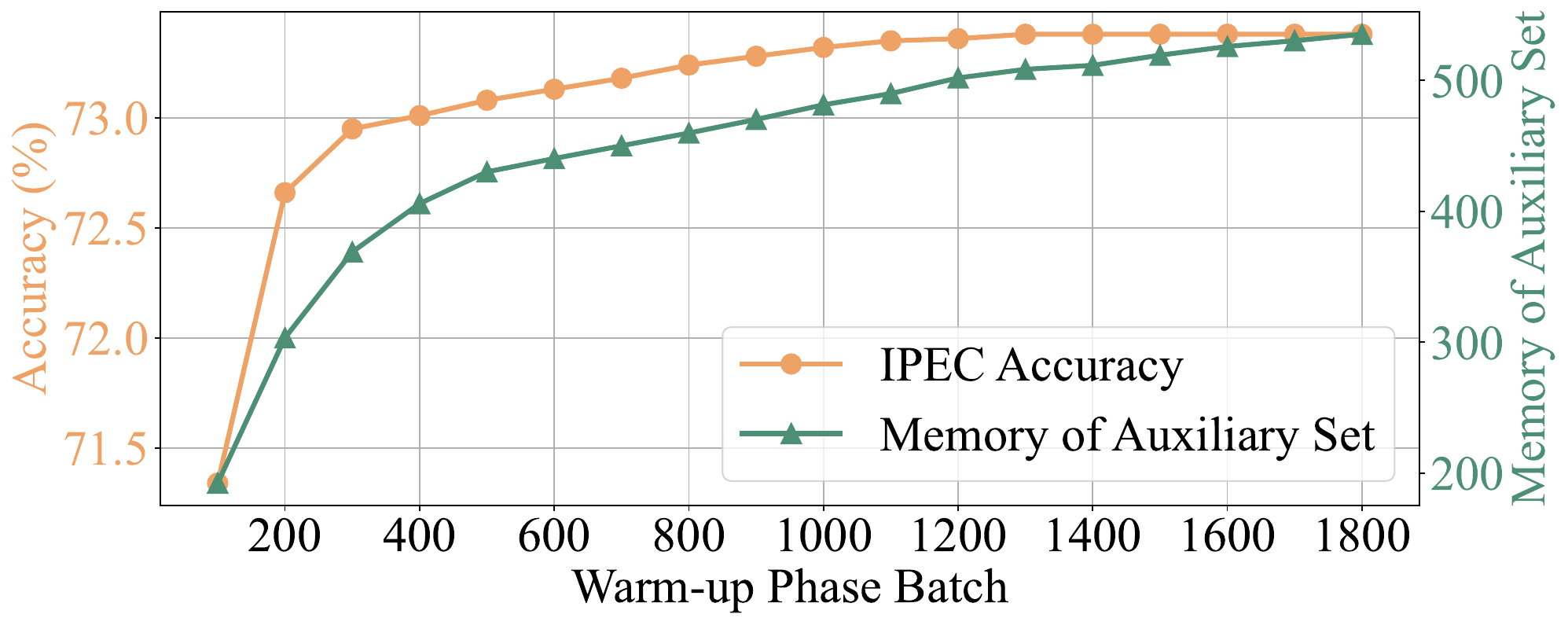}
  \caption{Warm-up Phase Effectiveness. Increasing the warm-up duration yields significant initial accuracy gains that later diminish, while memory cost grows steadily.}

  \label{fig:warmup_fc100}
\end{figure}

First, the impact of warm-up length is evaluated using the FC100 dataset. As shown in Fig.~\ref{fig:warmup_fc100}, the number of epochs in the warm-up phase was varied, and the final generalization accuracy alongside corresponding memory consumption were recorded. On the FC100 dataset, increasing warm-up epochs from 100 to 400 yields a significant accuracy boost. This rapid improvement phase corresponds to posterior concentration, where the auxiliary set quickly accumulates informative samples, substantially enhancing prototype quality and making the model's estimations more reliable. Beyond 400 epochs, accuracy gains plateau as the auxiliary set reaches semantic saturation, where new samples add redundancy rather than value. Meanwhile, memory consumption increases linearly, revealing a clear accuracy-cost trade-off. A duration of 1300 epochs is thus identified as a favorable balance, securing near-peak performance without incurring excessive memory overhead.

\begin{figure}[t]
  \centering
  \includegraphics[width=0.99\linewidth]{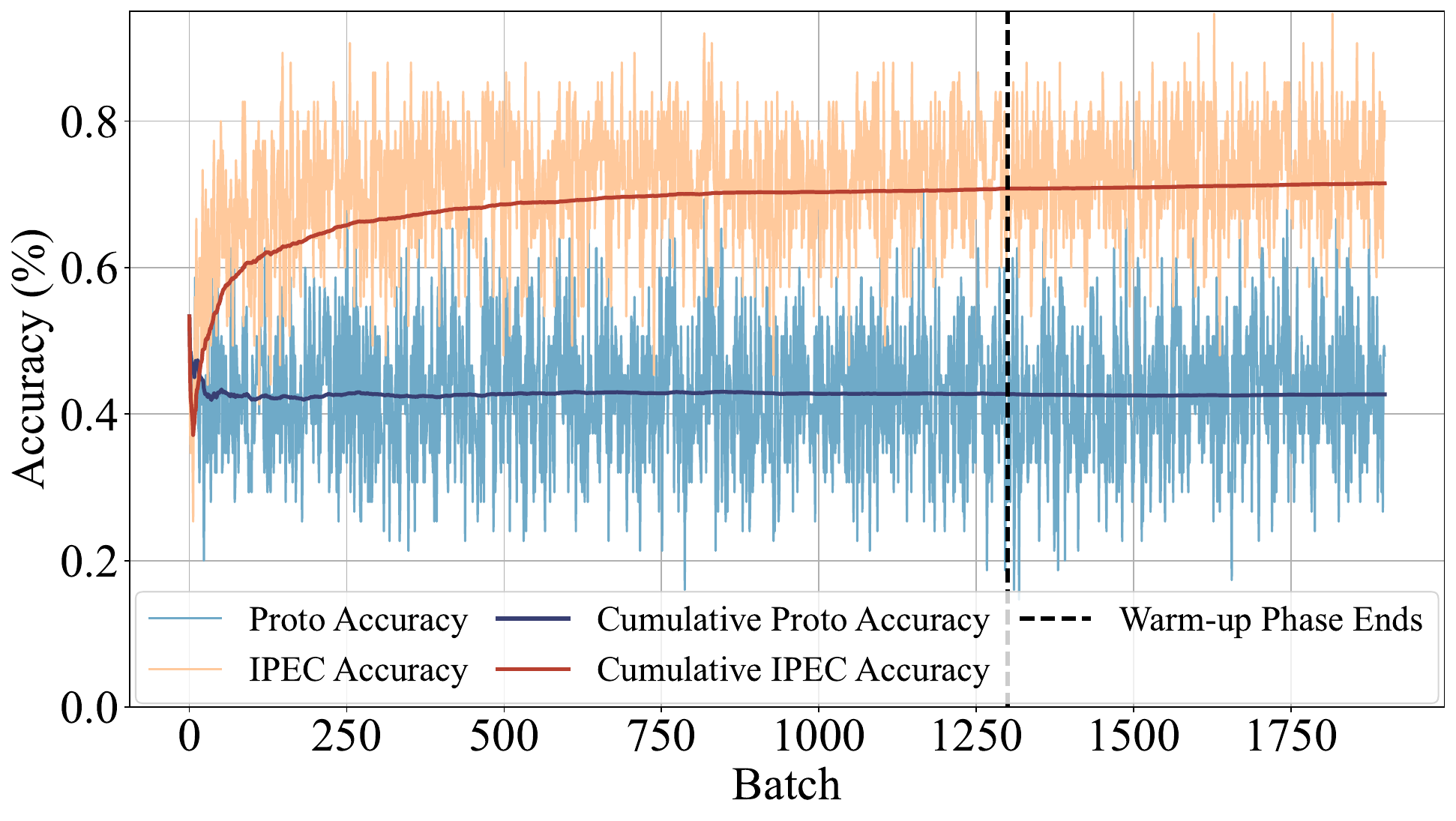}
  \caption{Learning Dynamics During Inference. IPEC's cumulative accuracy steadily improves and stabilizes during the warm-up phase as its auxiliary set matures, in contrast to the static accuracy of the PN baseline.}
  \label{fig:warmup_curve}
\end{figure}


\paragraph{The performance dynamics during and after warm-up}
\label{comparison}

To investigate the model’s behavior throughout the inference process, cumulative accuracy curves over batches are plotted in Fig.~\ref{fig:warmup_curve}. During the warm-up phase (up to batch 1300), cumulative accuracy steadily rises and stabilizes as the auxiliary set matures and initial batch-wise variance subsides. This trend confirms the creation of reliable, posterior-informed prototypes. Crucially, once the warm-up concludes and the set is fixed for the testing phase, both generalization and prototype accuracies remain high and stable.

In summary, the experiments confirm that the warm-up phase is critical for establishing a reliable auxiliary set. An appropriately chosen warm-up length allows the model to achieve high accuracy with manageable resource consumption. Once this phase is complete, the model can maintain stable and high performance without further updates, validating the design of the two-stage inference strategy.

\subsection{Mechanistic Analysis and Visualization}

\subsubsection{Confidence-Based Filtering: Theory and Threshold Optimization}

\paragraph{Rationale for Confidence Metric Selection}

To ensure that only the most reliable query samples are incorporated into the auxiliary set \( A_c \) for prototype refinement, it is essential to design a robust sample selection mechanism based on well-justified confidence metrics. We consider three principal candidates for evaluating prediction reliability:

Global Confidence via Entropy (\( \Delta \)): As defined in Eq. (\ref{Delta}), this metric captures the overall certainty of the model's predictive distribution. A low entropy (high \( \Delta \)) indicates that the model assigns a high probability to a single class, making it a strong indicator of global prediction quality.

Local Confidence via Logit Margin (\( \Delta' \)): Defined in Eq. (\ref{Delta'}), this metric quantifies the margin between the top predicted class and the second-highest scoring class. A large margin (high \( \Delta' \)) reflects the model's ability to clearly distinguish the predicted class from others, indicating strong local decision confidence.

Maximum Logit/Probability Confidence (Conf Max): This simple and intuitive measure corresponds to the highest probability (or logit score) assigned by the model. A high \( l_{\text{max}} \) is typically interpreted as strong confidence in the predicted class.

While each of these metrics captures a different perspective on prediction confidence, an effective sample selection strategy should ideally leverage metrics that are both informative and complementary. Metrics that are highly correlated tend to be redundant and may not contribute additional discriminative power, potentially complicating threshold tuning.

To investigate the relationships among these metrics, we compute the correlation coefficient (\( r^2 \)) between each pair of confidence measures across four benchmark datasets: Mini-ImageNet (Mini), Tiered-ImageNet (Tiered), CIFAR and FC100. The results are presented in Table~\ref{table:r2_threshold}.

\begin{table}
    \caption{Correlation coefficient values of different threshold type correlations across datasets}
    \label{table:r2_threshold}
    \centering
    \small
    \begin{tabular}{l|cccc}
    \toprule
    Threshold Comparison & Mini & Tiered & CIFAR & FC100 \\
    \midrule
    Conf Entropy VS Conf Max    & 0.79 & 0.79 & 0.90 & 0.83 \\
    Conf Diff VS Conf Max       & 0.71 & 0.68 & 0.74 & 0.70 \\
    Conf Diff VS Conf Entropy   & 0.52 & 0.50 & 0.55 & 0.48 \\
    \bottomrule
    \end{tabular}
\end{table}

The \( r^2 \) values reveal several insights. First, Conf Entropy (\( \Delta \)) and Conf Max exhibit high correlations (0.79–0.90), suggesting they capture similar aspects of confidence. Likewise, Conf Diff (\( \Delta' \)) and Conf Max also show strong correlations (0.68–0.74). By contrast, Conf Diff (\( \Delta' \)) and Conf Entropy (\( \Delta \)) demonstrate consistently lower correlations, with \( r^2 \) values ranging from 0.48 to 0.55. This weaker correlation implies that \( \Delta \) and \( \Delta' \) reflect distinct and complementary aspects of prediction certainty—global versus local confidence.

Therefore, \( \Delta \) (global entropy confidence) and \( \Delta' \) (local logit margin confidence) are adopted as our dual selection criteria, as formulated in Eq. (\ref{criteria}). The combination ensures that accepted samples are not only globally confident (high \( \Delta \)) but also locally well-separated from alternatives (high \( \Delta' \)), leading to a more refined and reliable auxiliary set.

\vspace{1em}

\paragraph{Enhanced Prediction Accuracy and Confidence with IPEC}

\begin{figure}[!t]
\centering
\subfloat[]{\includegraphics[width=0.49\linewidth]{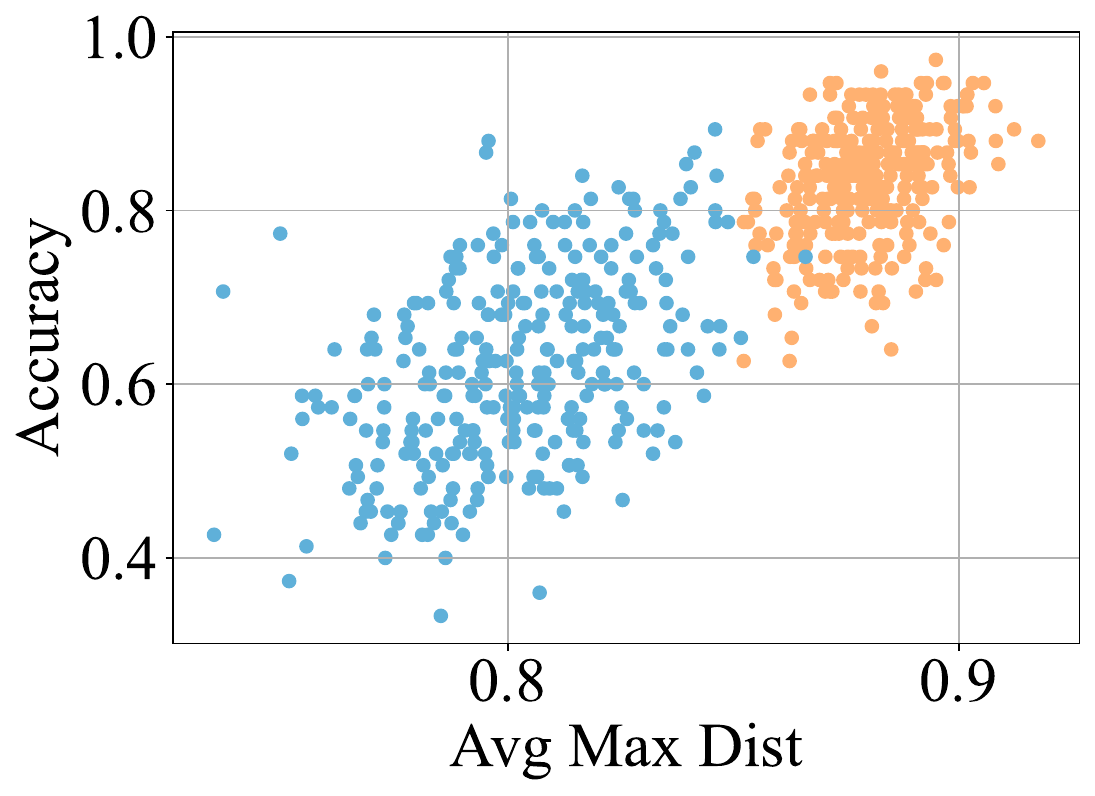}%
\label{fig:miniimagenet}}
\hfil 
\subfloat[]{\includegraphics[width=0.49\linewidth]{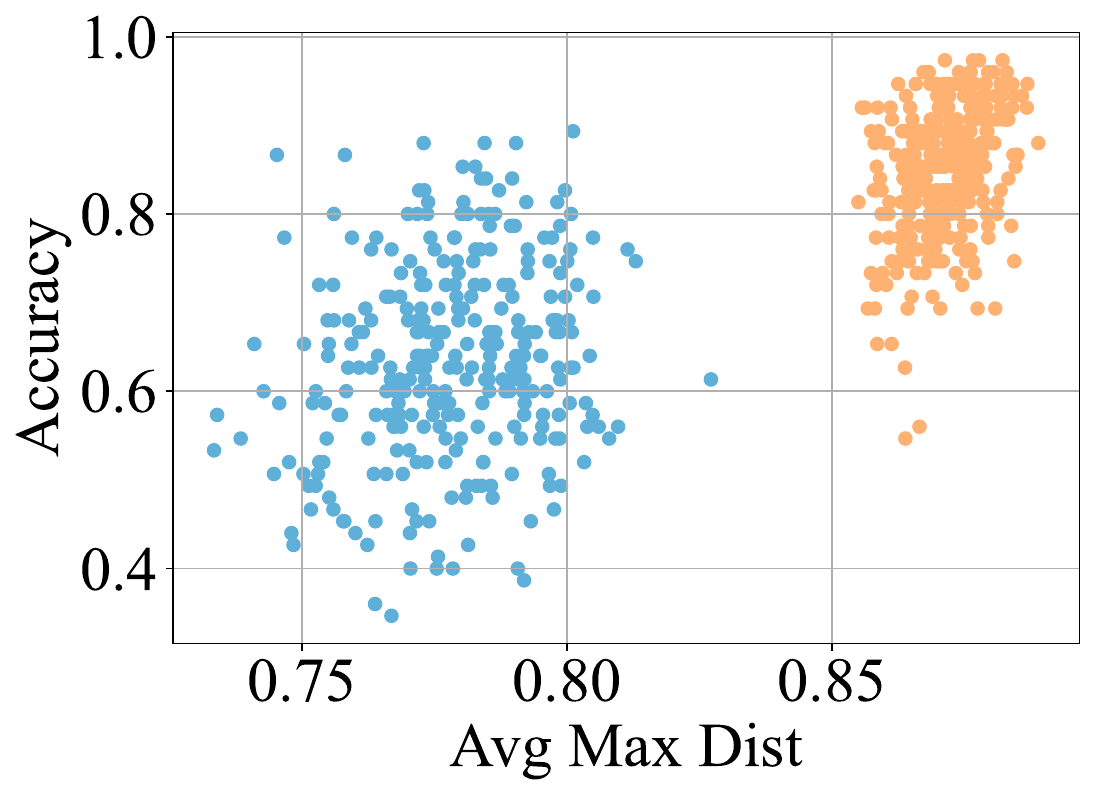}%
\label{fig:tieredimagenet}}


\subfloat[]{\includegraphics[width=0.49\linewidth]{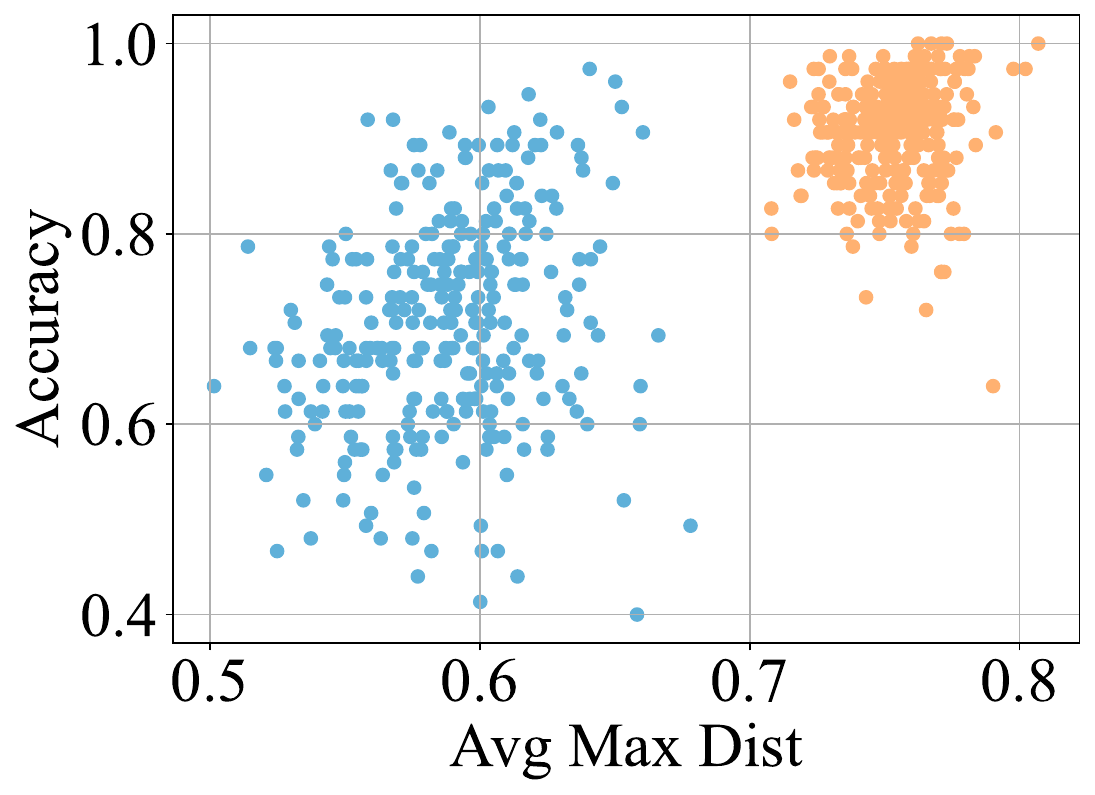}%
\label{fig:cifarfs}}
\hfil 
\subfloat[]{\includegraphics[width=0.49\linewidth]{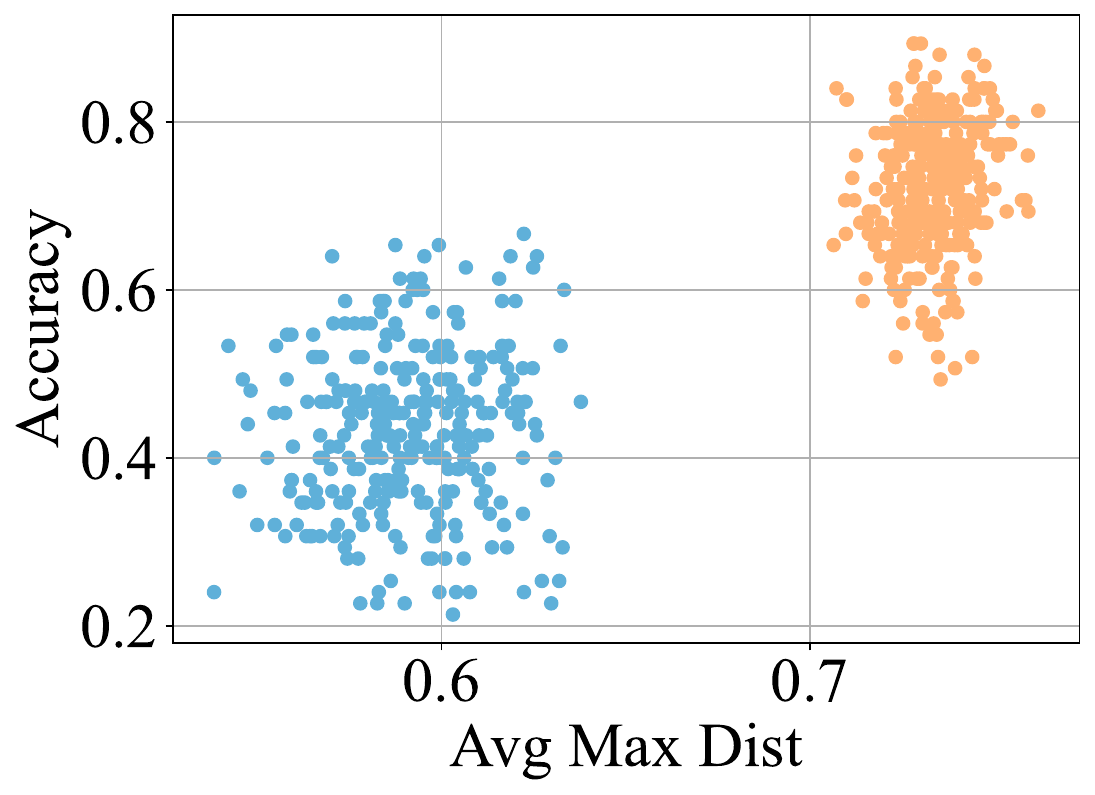}%
\label{fig:fc100}}

\caption{Scatter Plots. Showing the relationship between per-batch average accuracy and per-batch average maximum logit across four datasets. Each figure compares our IPEC method (blue dots) against a standard baseline PN (orange dots), which does not incorporate auxiliary set refinement. (a) MiniImageNet. (b) TieredImageNet. (c) CIFAR-FS. (d) FC100.}
\label{fig:scatter_comparison}
\end{figure}

After determining the optimal thresholds \(\tau\) and \(\tau'\) for dual-confidence sample selection, we proceed to evaluate the broader impact of IPEC framework on prediction quality, which we define by both classification accuracy and model confidence. Although our selection mechanism uses \( \Delta \) and \( \Delta' \), the final predictions are based on the highest logit. Therefore, we investigate whether IPEC's prototype refinement yields predictions that are not only more accurate but also more confident, as measured by their maximum logit values. Fig. \ref{fig:scatter_comparison} presents several consistent and compelling patterns emerge across all datasets:

\begin{itemize}
\item Clear Separation and Superiority of IPEC: IPEC’s data points form a well-separated cluster that consistently lies above and to the right of the PN cluster, demonstrating clear performance superiority.

\item Improved Accuracy: The upward position of the IPEC cluster indicates that it achieves significantly higher accuracy on average across test batches compared to the PN.

\item Increased Prediction Confidence: The rightward shift of the IPEC cluster reflects that its predictions have higher average maximum logit values. This means that the chosen class in IPEC predictions is assigned a higher score, indicating stronger model certainty.

\item Reduced Variance: Visually, the IPEC cluster exhibits a tighter grouping, particularly along the accuracy axis, suggesting more stable and reliable performance across different test batches.
\end{itemize}

The refinement of class prototypes within the IPEC framework concurrently enhances prediction accuracy and confidence. Such a synergistic improvement, where heightened confidence is a direct consequence of superior prototype quality, validates the framework's practical efficacy in advancing FSL.

\vspace{1em}

\subsubsection{Feature Space Visualization with t-SNE}

\begin{figure*}[!t]
\centering
\subfloat[]{\includegraphics[width=0.32\textwidth]{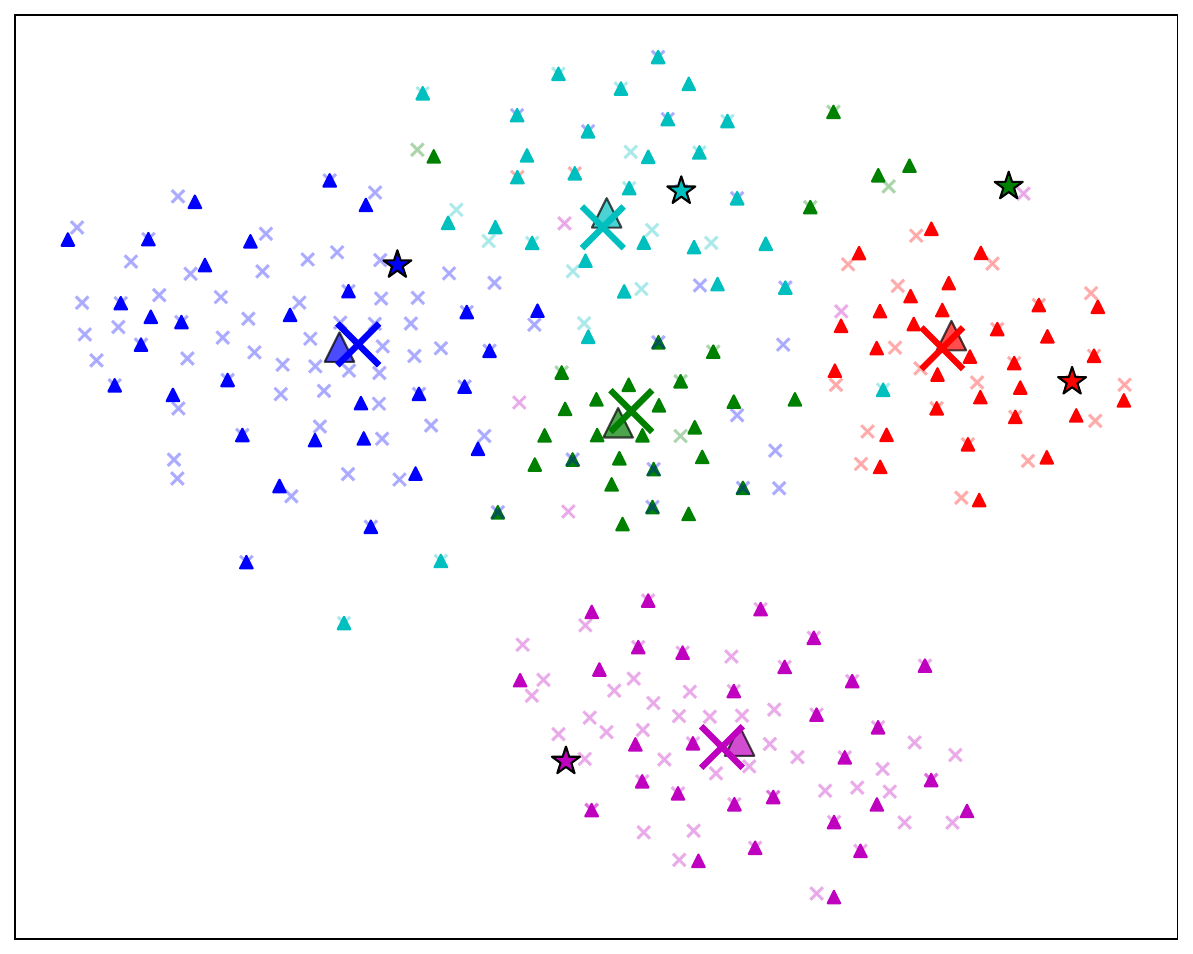}\label{subfig:tsne_cifar_initial}}%
\hfil
\subfloat[]{\includegraphics[width=0.32\textwidth]{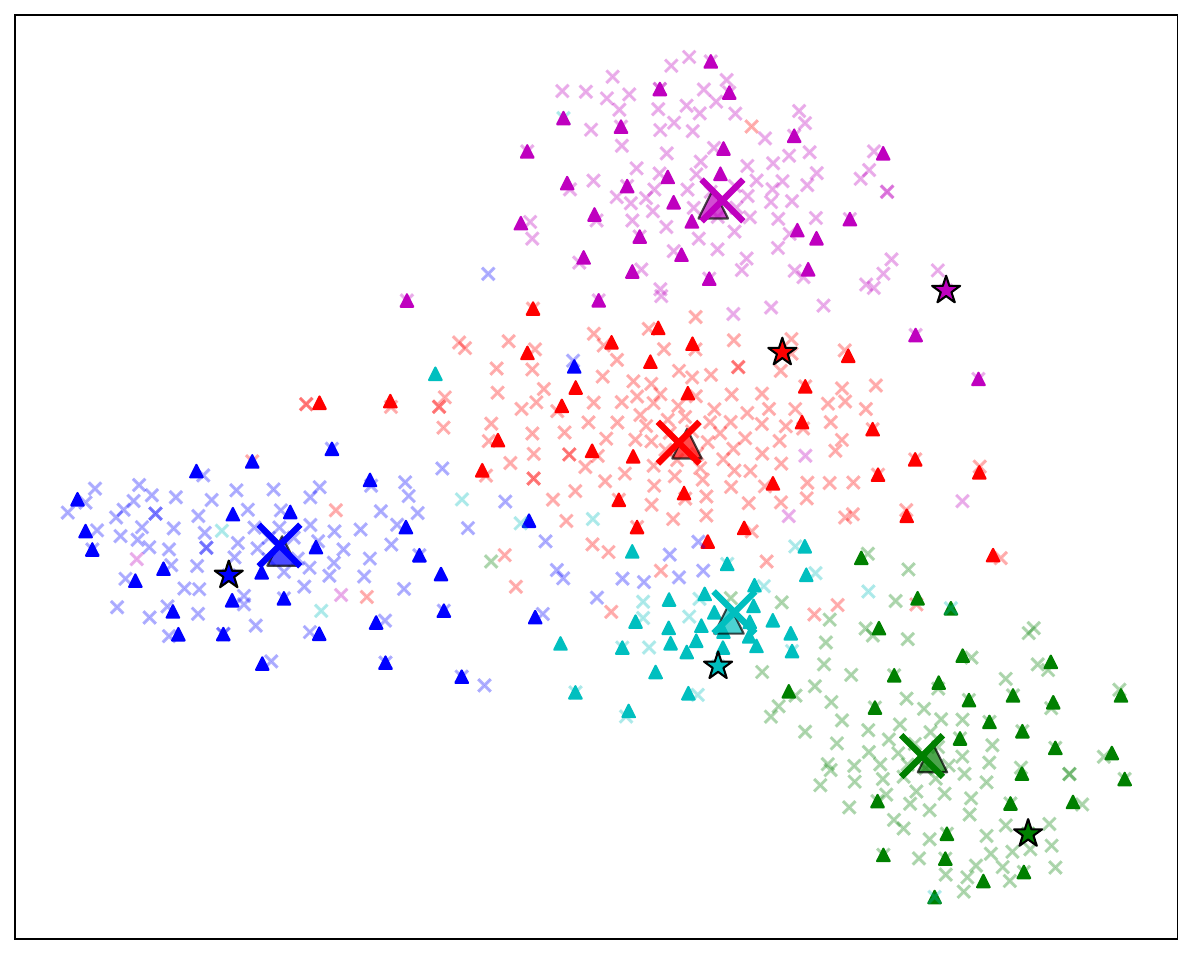}\label{subfig:tsne_cifar_middle}}%
\hfil
\subfloat[]{\includegraphics[width=0.32\textwidth]{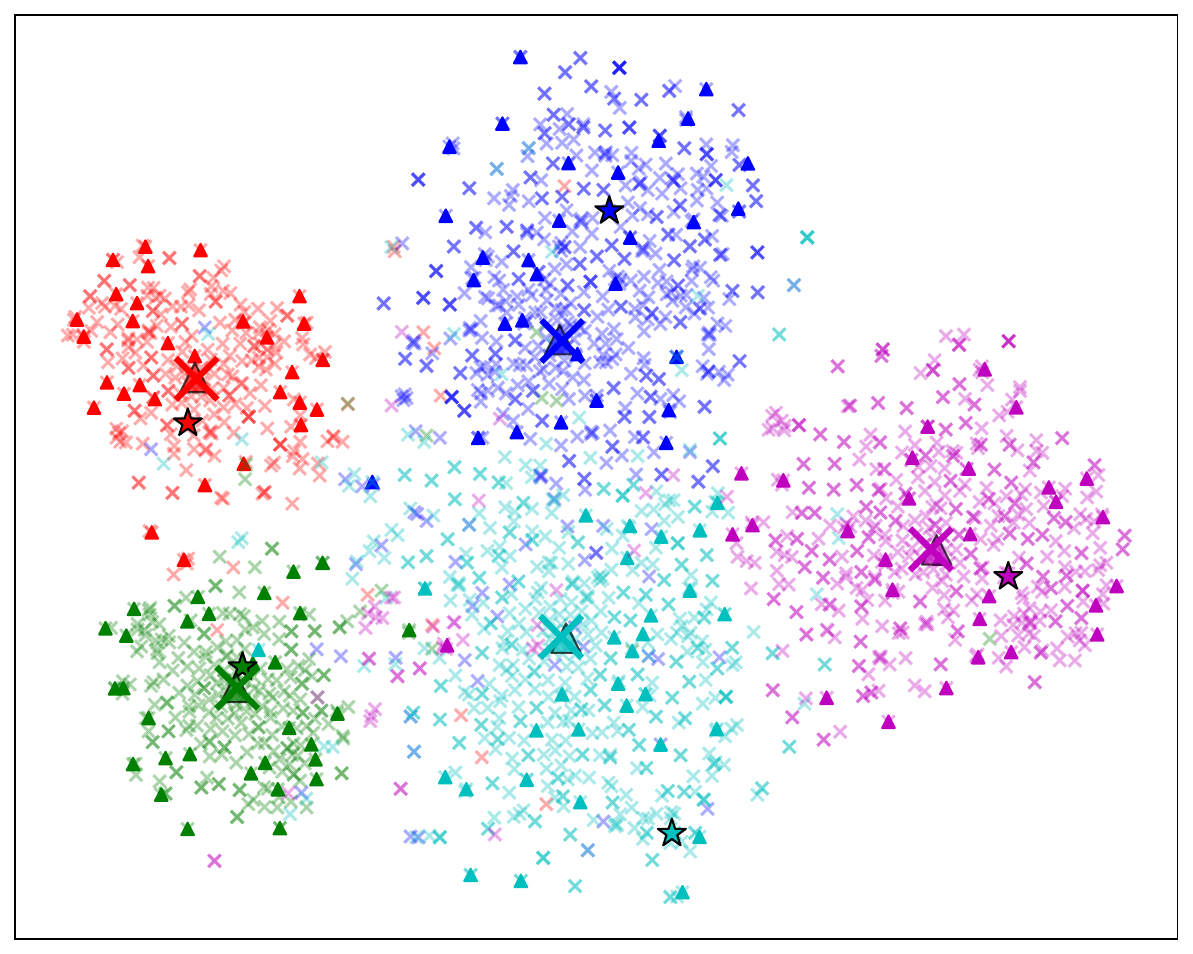}\label{subfig:tsne_cifar_late}}


\subfloat[]{\includegraphics[width=0.32\textwidth]{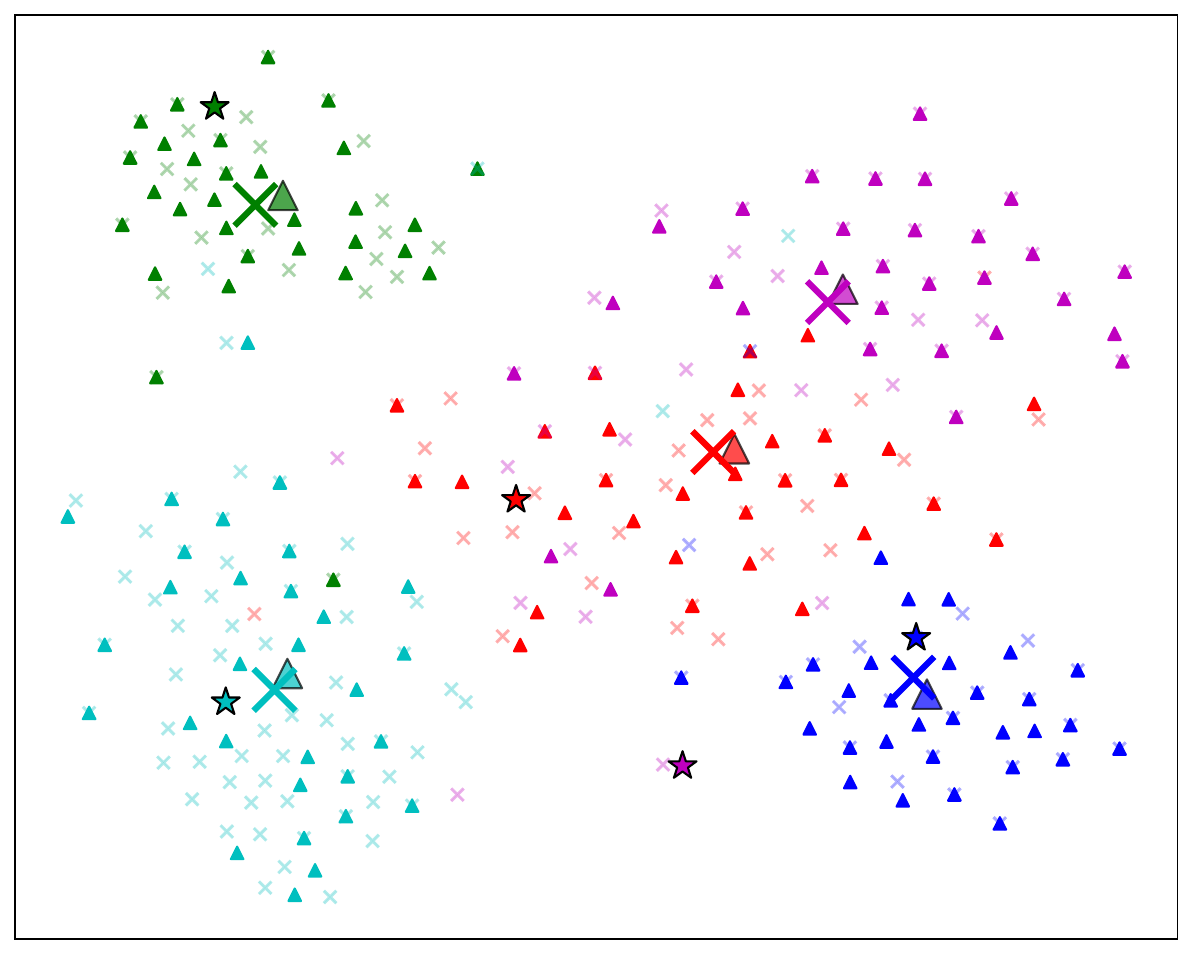}\label{subfig:tsne_mini_initial}}%
\hfil
\subfloat[]{\includegraphics[width=0.32\textwidth]{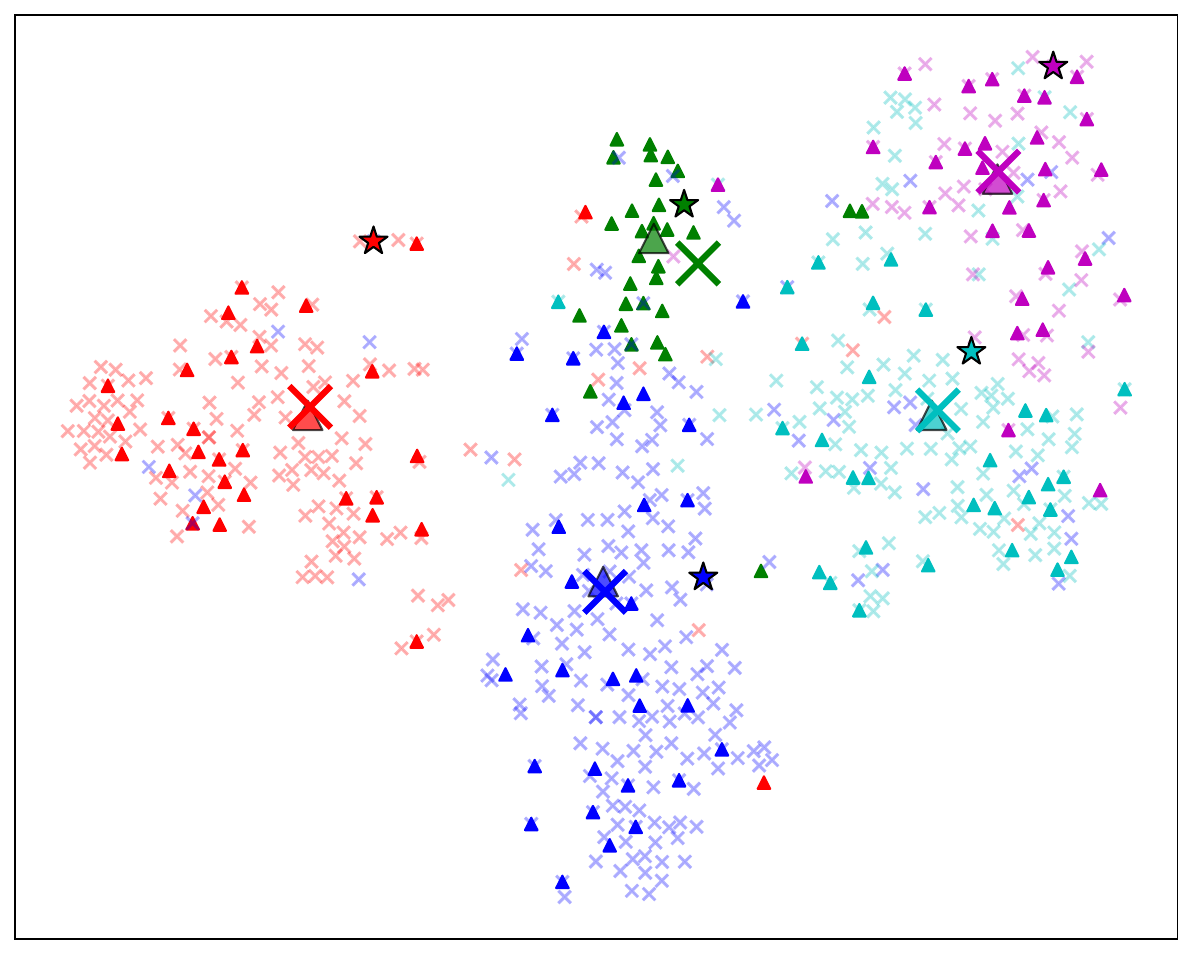}\label{subfig:tsne_mini_middle}}%
\hfil
\subfloat[]{\includegraphics[width=0.32\textwidth]{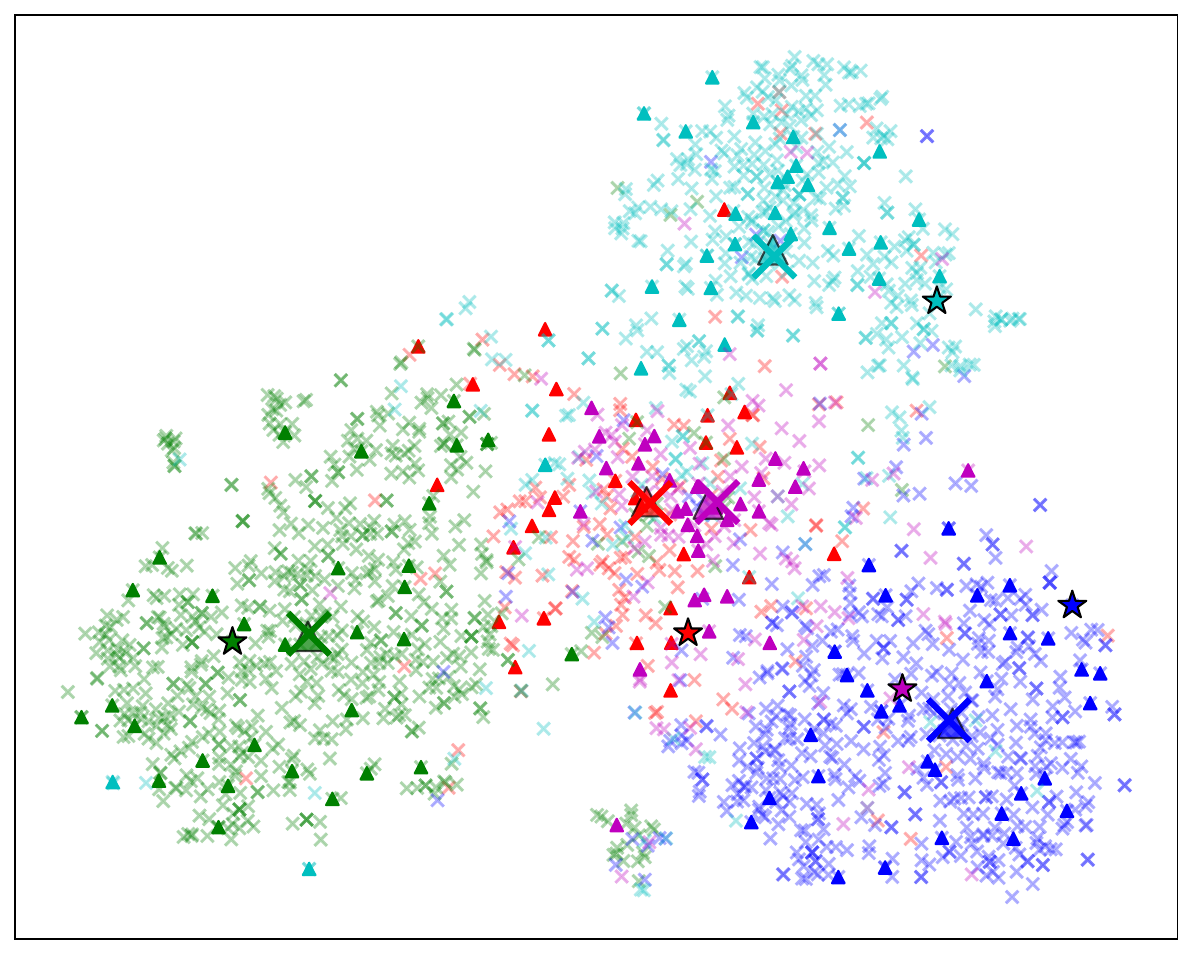}\label{subfig:tsne_mini_late}}

\caption{T-SNE Visualization of Features. The large triangle ($\mathbf{\blacktriangle}$) denotes the mean of query sample features for a given class (i.e., the query centroid), while the small triangles ($\blacktriangle$) represent individual query samples. The bold cross ($\boldsymbol{\times}$) signifies the mean of auxiliary sample features utilized in the current batch for that class (i.e., the auxiliary centroid), whereas the semi-transparent small crosses ($\times$) indicate the individual auxiliary samples drawn in that batch. The star ($\star$) represents the support sample for the class (which, in a 1-shot setting, also serves as the support centroid). Different colors distinguish different classes. (a)-(c) Feature evolution on CIFAR and (d)-(f) on MiniImageNet, showing initial, middle, and late warm-up stages (left to right).}
\label{fig:tsne}
\end{figure*}

To intuitively illustrate the distribution and evolution of Support, Query, and Auxiliary features during the warm-up phase, we employ the t-SNE technique. This is applied to visualize the low-dimensional embeddings of features extracted by the model at different training batches on both the CIFAR and MiniImageNet datasets. The experiments are conducted under a 5-way 1-shot task setting. The results are depicted in Fig. \ref{fig:tsne}, where the subplots from left to right represent the feature distributions of a specific batch during the initial, middle, and late stages of the warm-up phase, respectively.

Observations across the training stages reveal a distinct dynamic within the feature space. Initially, the auxiliary centroid is positioned nearer to its corresponding query centroid than the support sample. As training progresses, the auxiliary feature clusters for each class become increasingly compact and separable. This development is accompanied by the auxiliary centroid's progressive convergence toward the query centroid, a process that culminates in the late stage where it aligns very closely, or nearly coincides, with the query centroid. In contrast, the support sample consistently maintains a discernible distance from the query centroid throughout all stages. Concurrently, a visible increase in the number and density of auxiliary samples within each class is evident, indicating the expansion of the auxiliary set as the warm-up phase advances.

Concurrently with the expansion of the auxiliary set, the auxiliary centroid ($\boldsymbol{\times}$) demonstrates a stable convergence towards the query centroid ($\mathbf{\blacktriangle}$). This phenomenon is consistent with our theoretical analysis presented in Fig. \ref{fig:tradition} and Eq. (\ref{lln}). When the quantity of auxiliary samples is sufficiently large, their feature mean can more accurately estimate the true class centroid. Furthermore, because the auxiliary set is derived from a distribution representative of the query samples for that class, its mean provides an unbiased estimate of the expected query feature centroid. Consequently, the auxiliary and query centroids tend to converge.

In summary, the t-SNE visualizations robustly substantiate our theoretical assertions: (i) The auxiliary centroid is indeed closer to the query centroid than the support sample (particularly in scenarios with very few shots), thereby offering a more reliable class prototype. (ii) During the warm-up phase, as the number of training batches increases, the utilized auxiliary sample set progressively expands, leading the auxiliary centroid to stably converge towards the query centroid. This establishes a solid foundation for leveraging auxiliary information to enhance model generalization performance during the subsequent testing phase.

\section{Conclusions}
\label{sec:conclusion}

To address the challenge of data scarcity in large-scale testing, where conventional FSL methods are limited by processing each batch in isolation, a new approach is required. This paper introduces IPEC, a novel testing-time method that incrementally refines class prototypes by leveraging confidently classified samples from preceding test batches. This progressive mechanism reduces dependency on the initial support set and accumulates knowledge from the query stream. Extensive experiments demonstrate that IPEC achieves SOTA performance and, uniquely, improves its accuracy as more test data is processed, confirming its effectiveness for large-scale deployment.

Looking ahead, the focus of future work will be on two primary directions. First, we will develop more sophisticated update strategies for the auxiliary set to better represent the true data distribution. Second, we plan to validate IPEC on real-world industrial datasets to assess its practical robustness and efficiency, bridging the gap to industrial application.



\ifCLASSOPTIONcaptionsoff
  \newpage
\fi

\bibliographystyle{IEEEtran}

\bibliography{IEEEabrv,myref}

\end{document}